\documentclass{article}

\PassOptionsToPackage{compress}{natbib}
\usepackage[final]{neurips_2021}

\usepackage[utf8]{inputenc} 
\usepackage[T1]{fontenc}    
\usepackage{hyperref}       
\usepackage{url}            
\usepackage{booktabs}       
\usepackage{amsfonts}       
\usepackage{nicefrac}       
\usepackage{microtype}      
\usepackage{xcolor}         

\usepackage{graphicx}
\usepackage{subfigure}
\usepackage{enumitem}
\usepackage{mathtools}

\usepackage{amsmath}

\usepackage{bm}

\DeclareMathOperator*{\argmax}{arg\,max}

\newtheorem{innercustomgeneric}{\customgenericname}
\providecommand{\customgenericname}{}
\newcommand{\newcustomtheorem}[2]{%
  \newenvironment{#1}[1]
  {%
   \renewcommand\customgenericname{#2}%
   \renewcommand\theinnercustomgeneric{##1}%
   \innercustomgeneric
  }
  {\endinnercustomgeneric}
}

\newcustomtheorem{bootstrapping}{Bootstrapping}
\newcustomtheorem{datadist}{Data Distribution}
\newcustomtheorem{funcapprox}{Function Approximation}

\title{The Difficulty of Passive Learning\\in Deep Reinforcement Learning}

\author{%
  Georg Ostrovski \\
  DeepMind \\
  \texttt{ostrovski@deepmind.com} \\
  \And
  Pablo Samuel Castro \\
  Google Research, Brain Team \\
  \texttt{psc@google.com} \\
  \AND
  Will Dabney \\
  DeepMind \\
  \texttt{wdabney@deepmind.com} \
}

\begin{document}

\maketitle

\begin{abstract}
 Learning to act from observational data without active environmental interaction is a well-known challenge in Reinforcement Learning (RL). Recent approaches involve constraints on the learned policy or conservative updates, preventing strong deviations from the state-action distribution of the dataset. Although these methods are evaluated using non-linear function approximation, theoretical justifications are mostly limited to the tabular or linear cases. Given the impressive results of deep reinforcement learning, we argue for a need to more clearly understand the challenges in this setting.
 In the vein of Held \& Hein's classic 1963 experiment, we propose the ``tandem learning'' experimental paradigm which facilitates our empirical analysis of the difficulties in offline reinforcement learning. We identify function approximation in conjunction with fixed data distributions as the strongest factors, thereby extending but also challenging hypotheses stated in past work. Our results provide relevant insights for offline deep reinforcement learning, while also shedding new light on phenomena observed in the online case of learning control.
\end{abstract}

\section{Introduction}

Learning to act in an environment purely from observational data 
(i.e.~with no environment interaction), usually referred to as \textit{offline reinforcement learning}, has great practical as well as theoretical importance (see \citep{levine20offline} for a recent survey). 
In real-world settings like robotics and healthcare, it is motivated by the ambition to 
learn from existing datasets and the high cost of environment interaction. 
Its theoretical appeal
is that stationarity 
of the data distribution allows for more straightforward convergence 
analysis of learning algorithms. Moreover, decoupling learning from data generation
alleviates one of the major difficulties in the empirical analysis of common reinforcement learning agents,
allowing the targeted study of learning dynamics in isolation from their effects on behavior.

Recent work
has identified \textit{extrapolation error} as a major challenge for offline 
(deep) reinforcement learning 
\citep{achiam19towards,buckman21pessimism,fujimoto19offPolicy,fakoor2021continuous,liu2020provably,nair2020accelerating}, with \textit{bootstrapping} often highlighted as either the cause or an amplifier of the effect: 
The value of missing or under-represented state-action pairs
in the dataset can be over-estimated, either transiently (due to insufficient training or data)
or even asymptotically (due to modelling or dataset bias), resulting in a potentially severely
under-performing acquired policy.
The corrective feedback-loop \citep{kumar20discor},
whereby value over-estimation is self-correcting via \textit{exploitation} during interaction with the environment
(while under-estimation is corrected by \textit{exploration}),
is critically missing in the offline setting.

To mitigate this, typically one of a few related strategies are proposed: 
policy or learning update constraints preventing deviations
from states and actions well-covered by the dataset or satisfying certain uncertainty bounds
\citep{fujimoto19benchmarking,fujimoto19offPolicy,kumar2019,kumar2020,achiam19towards,wang2020critic,wu21uncertainty,nair2020accelerating,wu2019behavior,yu2020mopo}, 
pessimism bias to battle value over-estimation \citep{buckman21pessimism,kidambi20},
large and diverse datasets to improve state space coverage \citep{agarwal20optimistic}, or learned
models to fill in gaps with synthesized data \citep{schrittwieser2021online,matsushima2020deployment}. While many of these enjoy theoretical justification in the tabular or linear cases \citep{thomas2015high}, guarantees for the practically relevant non-linear case are mostly lacking.

\begin{figure}
    \centering
    \includegraphics[width=\textwidth]{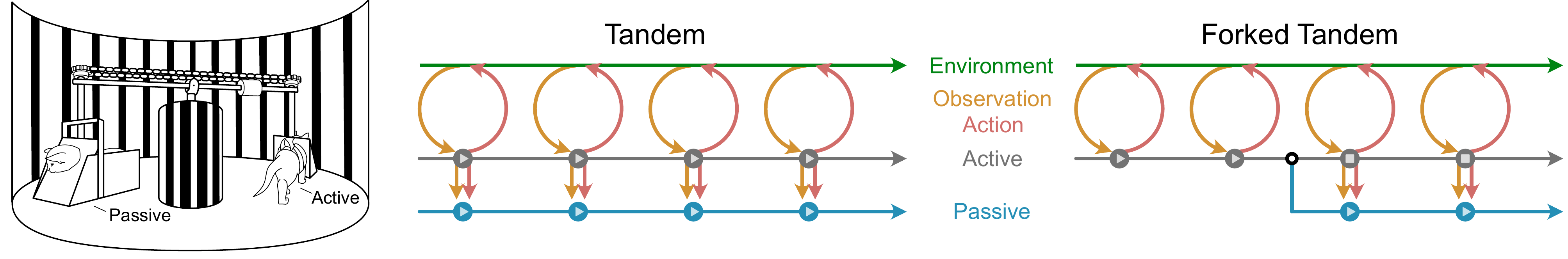}
    \includegraphics[width=\textwidth,trim = {0 0.4cm 0 0}, clip]{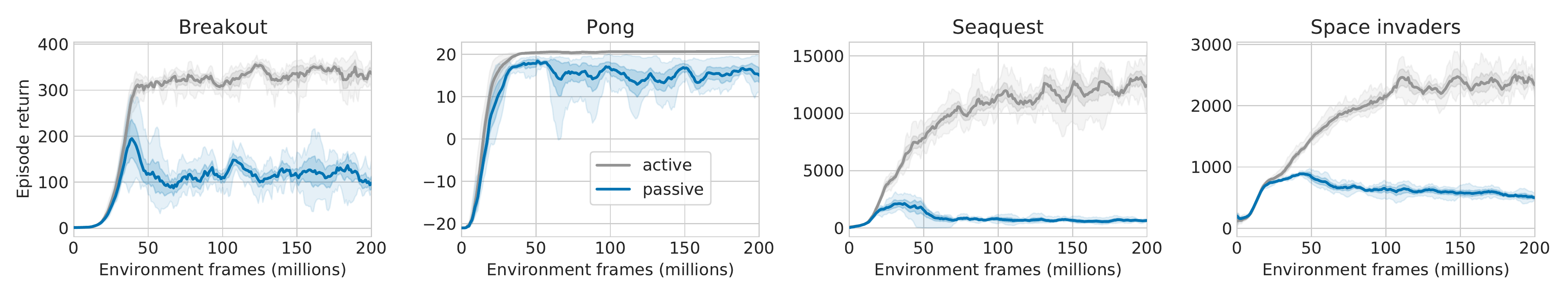}
    \caption{
    (\textbf{top-left}) \citet{held63movement} experiment setup.
    (\textbf{top-right}) Illustrations of the Tandem and Forked Tandem experiment setups.
    (\textbf{bottom}) Tandem (active and passive) performance on 4 selected Atari domains. In all figures, \textit{active} agent performance is shown in \color{gray}\textbf{gray}\color{black}.} 
    \label{fig:overview}
\end{figure}

In this paper we draw inspiration from the experimental paradigm introduced in the classic \citet{held63movement} experiment in psychology. The experiment involved coupling two young animal subjects' movements and visual perceptions to ensure that both receive the same stream of visual inputs, while only one can actively shape that stream by directing the pair's movements (Fig.~\ref{fig:overview}, top-left).
By showing that, despite identical visual experiences, only the actively moving subject acquired adequate visual acuity, the experiment established the importance of active locomotion in learning vision. 
Analogously, we introduce the `Tandem RL' setup, pairing an `active' and a `passive' agent in a
training loop where only the active agent drives data generation, while both perform
identical learning updates from the generated data\footnote{The `tandem' analogy is of course that of two riders, both of whom experience the same route, while only the front rider gets to decide on direction.}. 
By decoupling learning dynamics from its impact on data generation,
while preserving the non-stationarity of the online learning setting,
this experimental paradigm promises to be
a valuable analytic tool for the precise empirical study of RL algorithms. 

Holding architectures, losses, and crucially data distribution equal across the active and passive agents, or varying them in a controlled manner, we perform a detailed empirical analysis of the failure modes of passive (i.e.~non-interactive, offline)
learning, and pinpoint the contributing factors in properties of the data distribution,
function approximation and learning algorithm. Our study confirms some past intuitions for the failure
modes of offline learning, while refining and extending the findings in the deep RL case.
In particular, our results indicate an empirically less critical role for bootstrapping 
than previously hypothesized, 
while foregrounding erroneous extrapolation or over-generalization by a function approximator trained on an inadequate data distribution as the crucial challenge.
Among other things, our experiments draw a sharp boundary between the mostly well-behaved (and analytically well-studied) case of \textit{linear} function approximation, and the \textit{non-linear} case for which theoretical guarantees are lacking.
Moreover, we delineate different, more and less effective, ways of enhancing
the training data distribution to support successful offline learning, e.g. by
analysing the impact of dataset size and diversity, the stochasticity of the data generating policy,
or small amounts of self-generated data.
Our results provide hints towards a hypothesis relevant in both offline and online RL: 
robust learning of control with function approximation may require interactivity not 
merely as a data gathering mechanism, but as a counterbalance to a (sufficiently expressive) 
approximator's tendency to `exploit gaps' in an arbitrary fixed data distribution by excessive extrapolation.

\section{The Experimental Paradigm of Tandem Reinforcement Learning}

The Tandem RL setting, extending a similar analytic setup in \citep{fujimoto19offPolicy}, consists of two learning agents, one of which (the `active agent') performs the usual online training loop of interacting with an environment and learning from the generated data, while the other (the `passive agent') learns solely from data generated by the active agent, while only interacting with the environment for evaluation. We distinguish two experimental paradigms (see Fig.~\ref{fig:overview}, top-right):

\textbf{Tandem:} Active and passive agents start with independently initialized networks,
and train on an identical sequence of training batches in the exact same order.
    
\textbf{Forked Tandem:} An agent is trained for a fraction of its total training budget. It is then `forked' into active and passive agents, which start out with identical network weights. The active agent is `frozen', i.e.~receives no further training, but continues to generate data from its policy. The passive agent is trained on this generated data for the remainder of the training budget.

\subsection{Implementation}

Our basic experimental agent is `Tandem DQN', an active/passive pair of Double-DQN agents\footnote{Our choice of Double-DQN as a baseline is motivated by its relatively strong performance and robustness compared
to vanilla DQN \citep{dqn}, paired with its simplicity compared to later variants like Rainbow \citep{hessel2018rainbow}, which allows for more easily controlled experiments with fewer moving parts.} \citep{vanhasselt16deep}. Following the usual training protocol \citep{dqn}, the
total training budget is 200 iterations, each of which consists of 1M steps taken on the environment by the active agent, interspersed with regular learning updates (on one, or concurrently on both agents, depending on the paradigm), on batches of transitions sampled from the active agent's replay buffer. Both agents are independently evaluated on the environment for 500K steps after each training iteration. 

Most of our experiments are performed on the Atari domain \citep{bellemare13arcade}, using the exact algorithm and hyperparameters from \citep{vanhasselt16deep}. We use a fixed set of four representative games to demonstrate most of our empirical results, two of which (\textsc{Breakout}, \textsc{Pong}) can be thought of as easy and largely solved by baseline agents, while the others (\textsc{Seaquest}, \textsc{Space Invaders}) have non-trivial learning curves and remain challenging. Unless stated otherwise, all results 
show averages over at least 5 seeds, with confidence intervals indicating variation over seeds. In comparative plots, \textbf{boldface} entries indicate the default Tandem DQN configuration, and \color{gray}\textbf{gray }\color{black} lines always correspond to the active agent's performance.

\subsection{The Tandem Effect}

We begin by reproducing the striking observation in \citep{fujimoto19offPolicy} that the passive learner generally fails to
adequately learn from the very data stream that is demonstrably sufficient for its architecturally identical active counterpart; we refer to this phenomenon as the `tandem effect' (Fig.~\ref{fig:overview}, bottom). We ascertain the generality of this finding by replicating it across a broad suite of environments and agent architectures: Double-DQN on 57 Atari environments (Appendix Figs.~\ref{fig:atari57}~\&~\ref{fig:atari57_normalized}), adapted agent variants on four Classic Control domains from the OpenAI Gym library \citep{brockman2016openai}
 and the MinAtar domain \citep{young19minatar} (Appendix Figs.~\ref{fig:vanilla_tandem_cc}~\&~\ref{fig:minatar}), and the
distributed R2D2 agent \citep{kapturowski2018recurrent} (Appendix Fig.~\ref{fig:r2d2}). Details on agents and environments are provided in the Appendix\footnote{We provide two Tandem RL implementations: \href{https://github.com/deepmind/deepmind-research/tree/master/tandem_dqn}{https://github.com/deepmind/deepmind-research/\allowbreak tree/\allowbreak master/tandem\_dqn} based on the DQN Zoo \citep{dqnzoo2020github}, and \href{https://github.com/google/dopamine/tree/master/dopamine/labs/tandem_dqn}{https://github.com/\allowbreak google/ \allowbreak dopamine/tree/master/dopamine/labs/tandem\_dqn} based on the Dopamine library \citep{castro18dopamine}.}.

Empirically, we make the informal observation that while active and passive Q-networks tend to produce similar values
for typical state-action pairs under the active policy (where the action is the active Q-value
function's argmax for a given state), their values are less correlated for other (non-argmax) actions, and in
fact the active and passive greedy policies of a Tandem DQN tend to disagree in a large fraction of states under the behavior distribution (on average $> 75\%$ of states, after 100M steps of training, across 57 Atari games; see Appendix Fig.~\ref{fig:atari57_disagreement}).
Moreover, in a fraction ($\approx 12 / 57$) of Atari games, we observe the passive agent's network to strongly over-estimate
a fraction of state-action values, with the over-estimation growing as training progresses.

\section{Analysis of the Tandem Effect} 

In line with existing explanations \citep{levine20offline}, we propose that the tandem effect is 
primarily caused by extrapolation error when certain state-action pairs are under-represented in 
the active agent's behavior data. 
Specifically with $\varepsilon$-greedy policies,
even small over-estimation of the values of rarely seen actions can lead to sufficient behavior deviations 
to cause catastrophic under-performance. 

We further extend this hypothesis: in the context of \textit{deep} reinforcement learning
(i.e.~with non-linear function approximation), 
an inadequate data distribution can drive over-generalization \citep{bengio2020interference}, 
making such erroneous extrapolation likely. 
While the tandem effect can show up as learning inefficiency even in the tabular case \citep{xiao2021sample}, it proves
especially pernicious in the case of non-linear function approximation, 
where erroneous extrapolation can lead to errors not just on completely unseen, 
but also rarely seen data, and can persist in the infinite-sample limit.

Coalescing this view and past analyses of challenges in offline RL (e.g.~\citep{levine20offline,fujimoto19offPolicy,liu2020provably})
into the following three potential contributing factors in the tandem effect
provides a natural structure to our analysis:

\begin{bootstrapping}{(B)} \label{fac:bootstrapping}
The passive agent's bootstrapping from poorly estimated (in particular, over-estimated) values causes any initially small
mis-estimation to get amplified.
\end{bootstrapping}

\begin{datadist}{(D)} \label{fac:data}
Insufficient coverage of sub-optimal actions under the active agent's policy may lead to their mis-estimation by the passive agent.
In the case of over-estimation, this may lead to the passive agent's under-performance.
\end{datadist}

\begin{funcapprox}{(F)} \label{fac:func}
A non-linear function approximator used as a Q-value function may tend to wrongly extrapolate
the values of state-action pairs underrepresented in the active agent's
behavior distribution. This tendency can be inherent and persistent, in the sense of being independent of initialization and not being reduced with increased training on the same data distribution.
\end{funcapprox}

These proposed contributing factors are not at all mutually exclusive; they may interact in causing or exacerbating the tandem effect. Insufficient coverage of sub-optimal actions under the active agent's behavior distribution \ref{fac:data} may lead to insufficient constraint on the respective values, which allows for effects of erroneous extrapolation by a function approximator \ref{fac:func}. Where this results in over-estimation, the use of bootstrapping \ref{fac:bootstrapping} carries the potential to `pollute' even well-covered state-action pairs by propagating over-estimated values (especially via the $\max$ operator in the case of Q-learning). In the next sections we empirically study these three factors in isolation, to establish their actual roles and relative contributions to the overall
difficulty of passive learning.

\subsection{The Role of Bootstrapping}

One distinguishing feature of reinforcement learning as opposed to supervised learning is its frequent use of
learned quantities as preliminary optimization targets, most prominently in what is referred to as \textit{bootstrapping} in the 
widely used TD algorithms \citep{sutton1988learning}, 
where preliminary estimates of the value function are used as update targets. In the
Double-DQN algorithm these updates take the form
$Q(s,a) \leftarrow r + \gamma \bar Q  (s', \argmax_{a'} Q(s',a'))$,
where $Q$ denotes the parametric Q-value function, 
and $\bar Q$ is the
target network Q-value function, i.e.~a time-delayed copy of $Q$. 

\begin{figure}
    \centering
    \includegraphics[width=\textwidth,trim = {0 0.4cm 0 0.3cm}, clip]{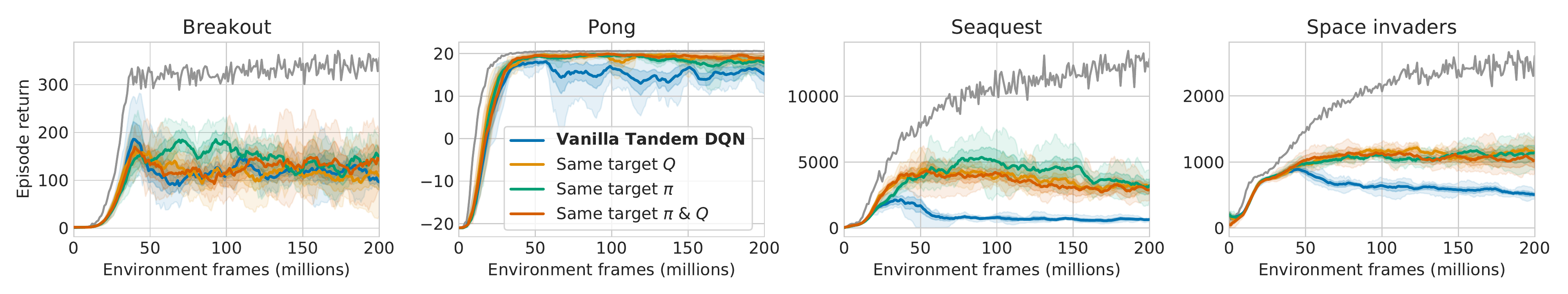}
    \caption{Active vs.~passive performance when using the active agent's target policy and/or value function
    for constructing passive bootstrapping targets.} 
    \label{fig:bootstrapping}
\end{figure}

Four value functions are involved in the active and passive updates of Tandem DQN: 
$Q_{\color{red} A}, \bar Q_{\color{red} A}, Q_{\color{blue} P}$ and $\bar Q_{\color{blue} P}$, where the ${\color{red} A}/{\color{blue} P}$ subscripts refer to the 
Q-value functions of the active and passive agents, respectively. 
The use of its own target network by the passive agent makes bootstrapping
a plausible strong contributor to the tandem effect.
To test this,
we replace the target values and/or policies in the update equation for the 
\emph{passive} agent, with the values provided by the \emph{active} agent's value functions:

\[
Q_P(s,a) \leftarrow \begin{cases}
r + \gamma \bar Q_{\color{blue} P}  (s', \argmax_{a'}  Q_{\color{blue} P}  (s',a'))\qquad \text{Vanilla Tandem DQN}\\
r + \gamma \bar Q_{\color{red} A} (s', \argmax_{a'}  Q_{\color{blue} P}  (s',a'))\qquad \text{Same Target $Q$} \\
r + \gamma \bar Q_{\color{blue} P} (s', \argmax_{a'}  Q_{\color{red} A}  (s',a'))\qquad \text{Same Target $\pi$} \\
r + \gamma \bar Q_{\color{red} A} (s', \argmax_{a'}  Q_{\color{red} A}  (s',a'))\qquad \text{Same Target $\pi$\&$Q$} \\
\end{cases}
\]

As shown in Fig.~\ref{fig:bootstrapping}, the use
of the active value functions as targets reduces the active-passive gap by only a small amount.
Note that when both active target values and policy are used, 
both networks are receiving \emph{an identical sequence of targets} for their update computations, a sequence that suffices for the active agent to learn a successful policy.
Strikingly, despite this the tandem effect appears largely preserved: 
in all but the easiest games (e.g.~\textsc{Pong}\footnote{\textsc{Pong} is an outlier in that it only has 3 actions, and in a large fraction of states actions have no (irreversible) consequences, making greedy policies somewhat robust to errors in the underlying value function.}) the passive agent fails to learn effectively.

To more precisely understand the effect of bootstrapping with respect to a potential value over-estimation by the passive agent, in Appendix Fig. \ref{fig:bootstrapping_overestimation} we also show the values of the passive networks in the above experiment compared to
those of the respective active networks. As hypothesised, we observe that the vanilla tandem setting leads to significant value over-estimation, and that indeed bootstrapping plays a substantial role in amplifying the effect: passive networks trained using the active network's bootstrap targets do not over-estimate compared to the active network at all.

These findings indicate that a simple notion of value over-estimation itself is not the fundamental cause of the tandem effect,
and that \textbf{\ref{fac:bootstrapping} plays an amplifying, rather than causal role}.
Additional evidence for this is provided below, 
where the tandem effect occurs in a purely supervised setting without bootstrapping.

\subsection{The Role of the Data Distribution}
\label{sec:distribution}

The critical role of the data distribution for offline learning is well established \citep{fujimoto19offPolicy,jacq19learning,liu2020provably,wang2021instabilities}. In particular, \citet{wang2020statistical} showed
that simpler notions of state-space coverage may not suffice for efficient offline learning with function approximation
(even in the linear case and under a strong realizability assumption); much stronger assumptions on the data distribution, 
typically not satisfied in practical scenarios, may actually be required.
Here we extend past analysis empirically, 
by investigating how properties of the data distribution (e.g.~stochasticity, stationarity, the size and diversity of the dataset,
and its proximity to the passive agent's own behavior distribution) affect its suitability for passive learning.

\begin{figure}
    \centering
    \includegraphics[width=\textwidth,trim = {0.2cm 0.2cm 0 0.3cm}, clip]{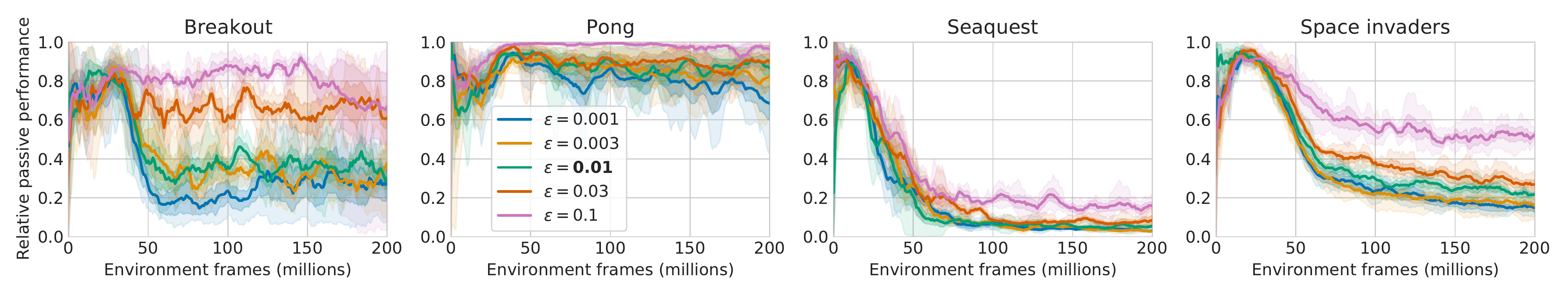}
    \includegraphics[width=\textwidth,trim = {0 0.4cm 0 0.3cm}, clip]{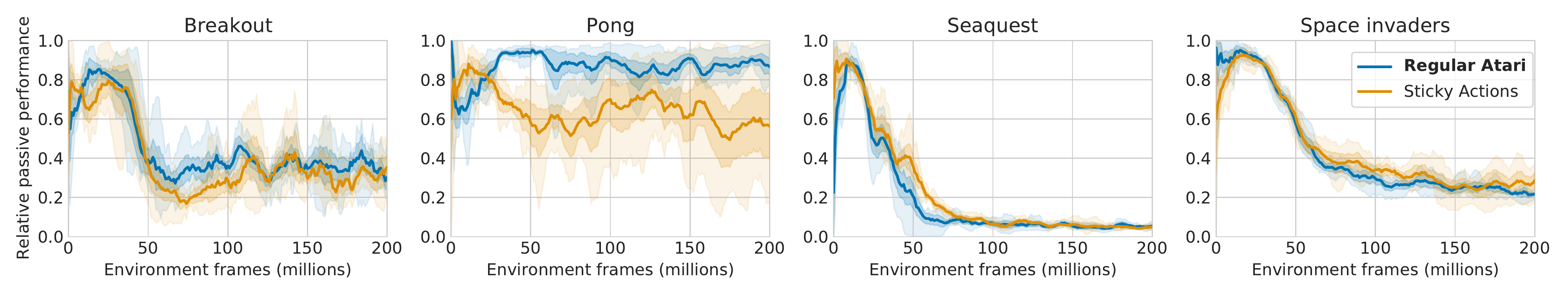}
    \caption{Passive as fraction of active performance for varying active $\varepsilon$-greedy behavior policies (\textbf{top}); regular Atari vs sticky-actions Atari (\textbf{bottom}). We report \textit{relative} passive performance, as active performance varies across configurations. See Appendix Figs.~\ref{fig:eps_return}~\&~\ref{fig:sticky_actions_return} for absolute performance.}
    \label{fig:stochasticity}
\end{figure}

\paragraph{The exploration parameter $\bm{\varepsilon}$}
A simple way to affect the data distribution's state-action coverage (albeit in a blunt and uniform way) is by varying
the exploration parameter $\varepsilon$ of the active agent's $\varepsilon$-greedy behavior policy (for training, not for evaluation). 
Note that a higher $\varepsilon$ parameter affects the \textit{active} agent's own
training performance, as its ability to navigate environments requiring precise 
control is reduced. In Fig.~\ref{fig:stochasticity} (top) we therefore report the 
\textit{relative} passive performance (i.e.~as a fraction
of the active agent's performance, which itself also varies across parameters), with absolute performance plots included in the Appendix for completeness (Fig.~\ref{fig:eps_return}). 
We observe that the relative passive performance
is indeed substantially improved when the active behavior policy's stochasticity (and as a consequence its coverage of
non-argmax actions along trajectories) is increased, and conversely it reduces with a greedier behavior policy, \textbf{providing evidence for the role of \ref{fac:data}}.

\paragraph{Sticky actions}
An alternative potential source of stochasticity is the environment itself, e.g.~the
 use of `sticky actions' in Atari \citep{machado2018revisiting}: with fixed probability, an agent action is ignored (and the previous action repeated instead). This type of environment-side stochasticity should not be expected to cause new actions to appear in the behavior data, and indeed 
Fig.~\ref{fig:stochasticity} (bottom) shows no substantial impact on the tandem effect.

\begin{figure}
    \centering
    \includegraphics[width=\textwidth,trim = {0 0.3cm 0 0.3cm}, clip]{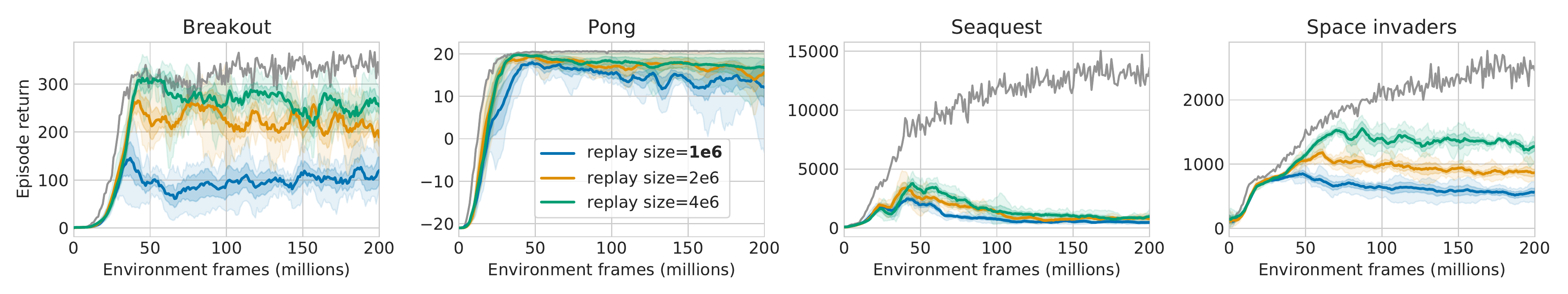}
    \caption{Active vs.~passive performance for different replay sizes (for passive agent).}
    \label{fig:size}
\end{figure}

\paragraph{Replay size}
Our results contrast with the strong offline RL results in \citep{agarwal20optimistic}.
We hypothesize that the difference is due to the vastly different dataset size (full training of 200M transitions vs.~replay buffer of 1M).
Interpolating between the tandem and the offline RL setting from \citep{agarwal20optimistic}, 
we increase the replay buffer size, thereby giving the passive agent access to somewhat larger data diversity 
and state-action coverage (this does not affect the active agent's training as the active agent is constrained to only sample from the
most recent 1M replay samples, as in the baseline variant). A larger
replay buffer somewhat mitigates the passive agent's under-performance (Fig.~\ref{fig:size}), though it appears to mostly slow down
rather than prevent the passive agent from eventually under-performing its active counterpart substantially.
As we suspect that a sufficient replay buffer size may depend on the effective state-space size of an environment,
we also perform analogous experiments on the (much smaller) classic control domains; results
 (Appendix Fig.~\ref{fig:size_cc}) remain qualitatively the same.
 
Note that for a fixed dataset size, sample diversity can take different forms. Many samples from a single policy may provide better coverage of states on, or near, policy-typical trajectories. Meanwhile, a larger collection of policies, with fewer samples per policy, provides better coverage of many trajectories at the expense of lesser coverage of small deviations from each.
To disentangle the impact of these modalities, while also shedding light on the role of
stationarity of the distribution, we next switch to the `Forked Tandem' 
variation of the experimental paradigm. 

\begin{figure}
    \centering
    \includegraphics[width=\textwidth,trim = {0 0.3cm 0 0.3cm}, clip]{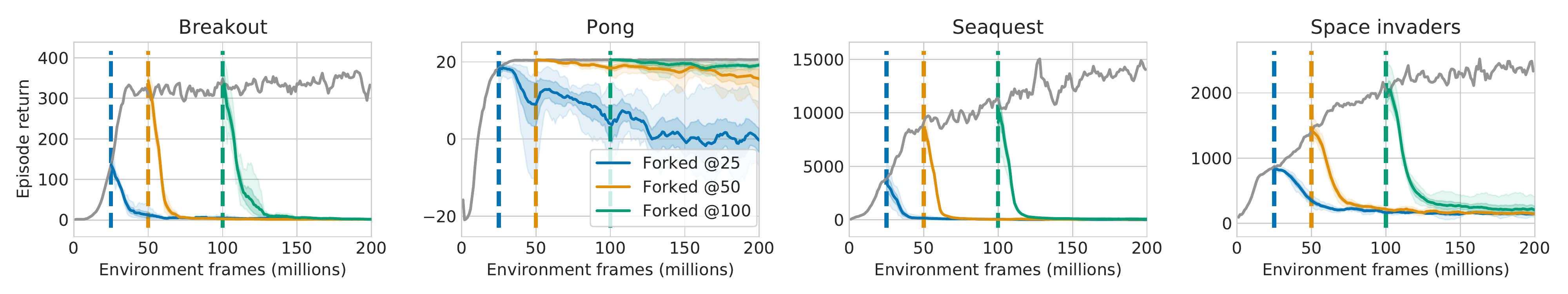}
    \includegraphics[width=\textwidth,trim = {0 0.3cm 0 0.3cm}, clip]{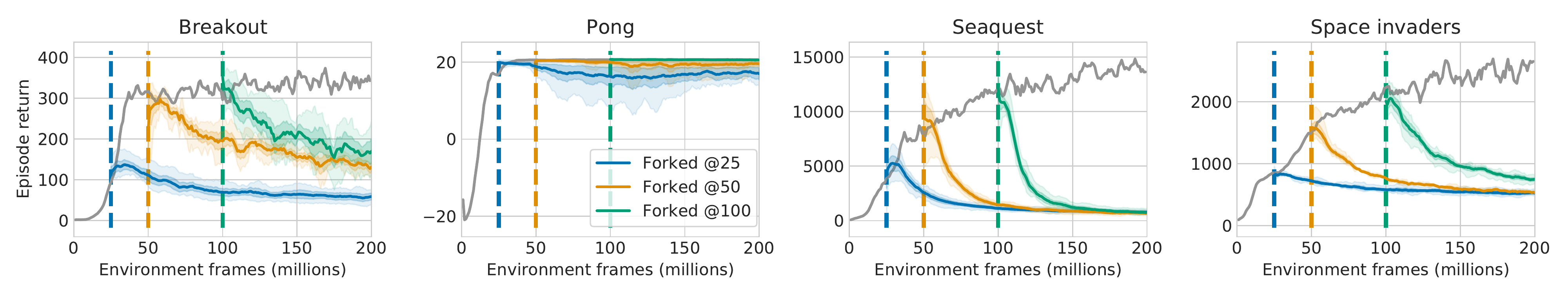}
    \caption{Performance of a Forked Tandem DQN, training passively after freezing its data generating policy (\textbf{top}) or its replay buffer (\textbf{bottom}). Vertical lines indicate forking time points.}
    \label{fig:forking}
\end{figure}

\paragraph{Fixed policy}
Upon forking, the frozen active policy is executed to produce training data for the passive agent, 
which begins its training initialized with the active network's weights.
Note that this constitutes a stronger variant of the tandem experiment.
At the time of forking, the agents do not merely share analogous
architectures and equal `data history', but also identical network weights (whereas
in the simple tandem setting, the agents were distinguished by independently initialized networks).
Moreover, the data used for passive training can be thought of as a `best-case scenario': 
generated by a single fixed policy, identical to the passive policy at the beginning of passive training.
Strikingly, the tandem effect is not only preserved but even exacerbated in this setting (Fig.~\ref{fig:forking}, top):
after forking, passive performance decays rapidly in all but the easiest games,
despite starting from a near-optimally performing initialization. 
This re-frames the tandem effect as not merely the difficulty of passively \emph{learning to act}, but even to
passively \emph{maintain performance}. Instability appears to be inherent in the very learning process itself, providing strong support to the hypothesis that an \textbf{interplay between \ref{fac:data} and \ref{fac:func} is critical to the tandem effect}.

In Appendix Fig.~\ref{fig:fork_eps_return} we additionally show that similarly to the regular tandem setting,
stochasticity of the active policy \textit{after forking} influences the passive agent's ability to
maintain performance.

\paragraph{Fixed replay}
A variation on the above experiments is to freeze the \textit{replay buffer} while continuing to train the passive policy from this fixed
dataset. Instead of a stream of samples from a single policy, this fixed data distribution now contains a fixed number of samples from a training process of the length of the replay buffer, i.e.~from a number of different policies. The collapse of passive performance here (Fig.~\ref{fig:forking}, bottom) is less rapid, yet qualitatively similar.
In Appendix Fig.~\ref{fig:ghd} we present yet another variant of this experiment with similar results, showing that the effect is robust to minor variations in the exact way of fixing the data distribution of a learning agent.

These experiments provide \textbf{strong evidence for the importance of \ref{fac:data}}: a larger replay buffer, containing samples from more diverse policies, can be expected to provide an improved
coverage of (currently) non-greedy actions, reducing the tandem effect. While the forked tandem begins passive learning with the seemingly advantageous high-performing initialization, state-action coverage is critically limited in this case. In the frozen-policy case, a large number of samples from the very same $\varepsilon$-greedy policy can be expected to provide very little coverage of non-greedy actions, while in the frozen-replay case, a smaller number of samples from multiple policies can be expected to only do somewhat better in this regard. In both cases the tandem effect is highly pronounced.

\begin{figure}
    \centering
    \includegraphics[width=\textwidth]{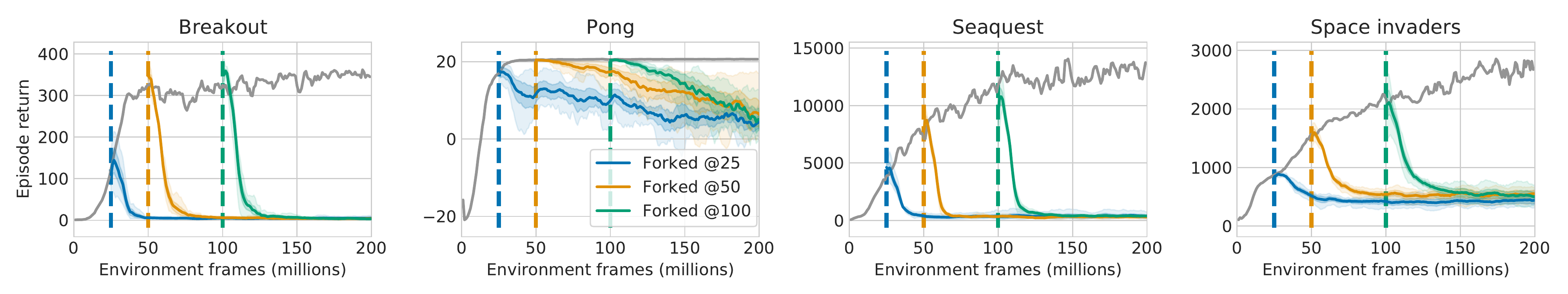}
    \caption{Passive performance in Forked Tandem DQN after policy evaluation with SARSA.} 
    \label{fig:sarsa}
\end{figure}

\paragraph{On-policy evaluation} The strength of the last two experiments lies in the observation that, since active and passive networks have identical parameter values at the beginning of passive training, 
their divergence cannot be attributed to small initial differences
getting amplified by training on an inadequate data distribution.
With so many factors held fixed, the collapse of passive performance when trained on the very data distribution produced by \textit{its own initial policy} begs the question whether off-policy Q-learning itself is to blame
for this failure mode, e.g.~via statistical over-estimation bias introduced by the $\max$ operator \citep{hasselt2010}.
Here we provide a negative answer, by performing on-policy evaluation with SARSA \citep{rummery1994line} (Fig.~\ref{fig:sarsa}), and even purely supervised regression on the Monte-Carlo returns (Appendix Fig.~\ref{fig:mc}), in the forked tandem setup. While evaluation succeeds, in the sense of minimizing evaluation error on the given behavior distribution, atypical action values under the behavior policy suffer substantial estimation error, resulting in occasional over-estimation. The resulting $\varepsilon$-greedy control policy under-performs the initial policy at forking time as catastrophically as in the other forked tandem experiments (more details in Appendix \ref{sec:eval}). 
\textbf{Strengthening the roles of \ref{fac:data} and \ref{fac:func} while further weakening that of \ref{fac:bootstrapping}}, these observations point to an inherent instability of offline learning, different from that of Baird's famous example \citep{baird95residual} or the `Deadly Triad' \citep{sutton2018reinforcement,van2018deep}; an instability that results purely from erroneous extrapolation by the function approximator, when the utilized data distribution does not provide adequate coverage of relevant state-action pairs.

\begin{figure}
    \centering
    \includegraphics[width=\textwidth,trim = {0 0.3cm 0 0.3cm}, clip]{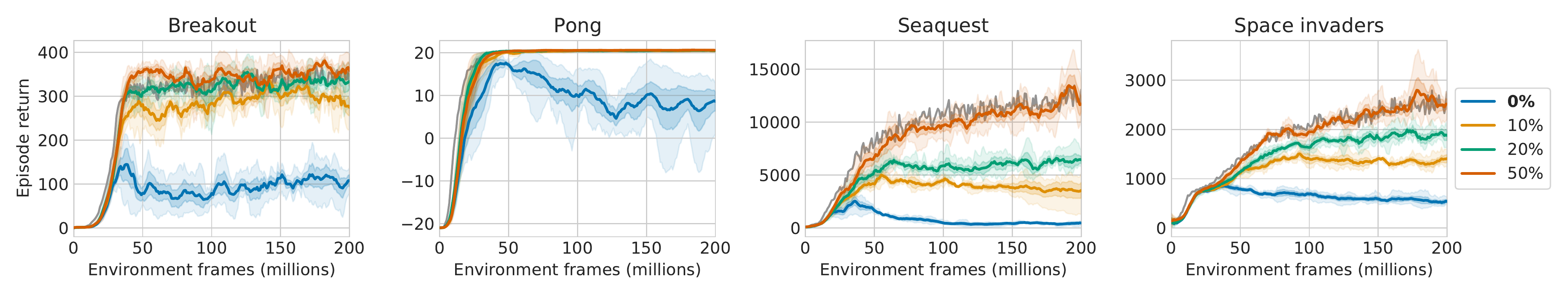}
    \caption{Passive performance for different amounts of self-generated data in the passive agent's replay batches.}
    \label{fig:own_data}
\end{figure}

\paragraph{Self-generated data}
Our final empirical question in this section is `How much data \textit{generated by the passive agent} is
needed to correct for the tandem effect?'. While a full investigation of this question exceeds the scope of this paper and is left for future work, the tandem setup lends itself to a simple experiment: \textit{both} agents interact with the environment and fill individual
replay buffers, one of them (for simplicity still referred to as `passive') however learns from data stochastically mixed
from both replay buffers. Fig.~\ref{fig:own_data} shows that even a moderate amount (10\%-20\%) of `own' data yields a substantial reduction of the tandem effect, while a 50/50 mixture completely eliminates it.

\subsection{The Role of Function Approximation} \label{sec:func}

We structure our investigation of the role of function approximation in the tandem effect into two categories:
the {\em optimization} process and the {\em function class} used.

\paragraph{Optimization} \citet{agarwal20optimistic} and \citet{obando2020revisiting} demonstrated that the Adam optimization algorithm \citep{kingma2014adam} outperforms RMSProp \citep{tieleman2012lecture} used in our experiments.
In Appendix Fig.~\ref{fig:optim} we show that while both active and passive agents perform better with Adam, the tandem effect itself is unaffected by the choice of optimizer.

Another plausible hypothesis is that the passive network suffers from under-fitting and requires more updates on the
same data to attain comparable performance to the active learner. 
Varying the number of passive agent updates per active agent update step,
we find that more updates \textit{worsen} the performance of the passive agent (Appendix Fig.~\ref{fig:replay_ratio}).
This rejects insufficient training as a possible cause, and further \textbf{supports the role of \ref{fac:data}}. 
We also note that, 
together with the forked tandem experiments in the previous section, this finding distinguishes the tandem effect from
the issue of estimation error in the offline learning setting \citep{xiao2021sample}: while in the tabular setting estimation error dominates
the learning challenge and a sufficient training duration (assuming full state-space coverage) guarantees convergence to a good solution,
this is not necessarily the case with function approximation trained on a given data distribution.

\paragraph{Function class} 
Given that the active and passive agents share an identical network architecture, the passive agent's under-performance
cannot be explained by an insufficiently expressive function approximator. Performing the tandem experiment with pure regression of the passive network's outputs towards the active network's (a variant of \textit{network distillation} \citep{distillation}), instead of TD based training, we observe that the performance gap is indeed vastly reduced and in some games closed entirely (see Appendix Fig.~\ref{fig:regression}); however, strikingly, it remains in some.

\begin{figure}
    \centering
    \includegraphics[width=0.7\textwidth,trim = {0 0.4cm 0 0.3cm}, clip]{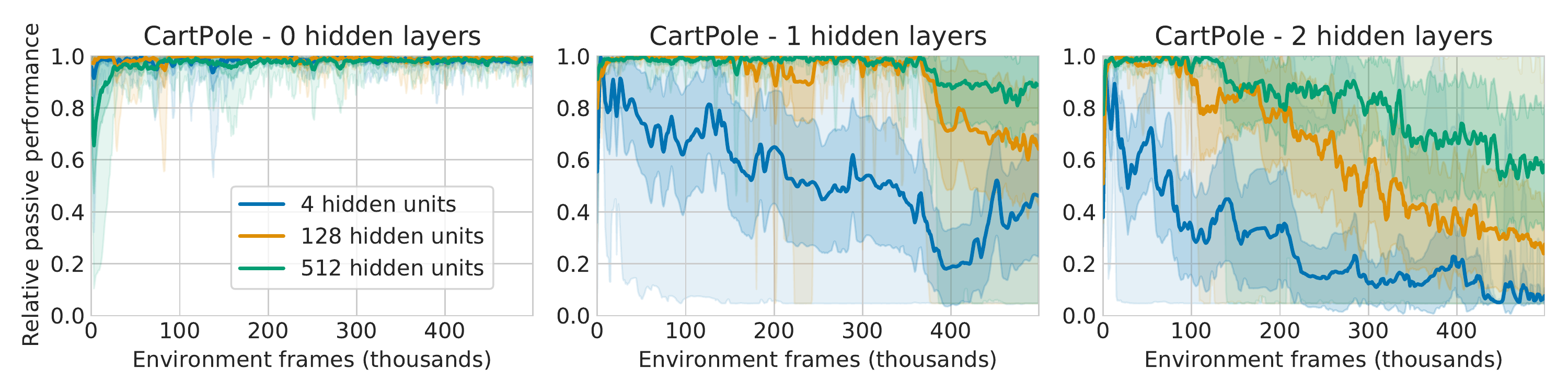}
    \caption{Passive performance as a fraction of active performance in CartPole: varying number of hidden layers and units.} 
    \label{fig:netarchCartpole}
\end{figure}

Next, we vary the function class of both networks by varying the depth and width of the utilized Q-networks on a set of Classic Control tasks. As can be seen in Fig.~\ref{fig:netarchCartpole} (and Appendix Fig.~\ref{fig:netarchAll}), 
the magnitude of the active-passive performance gap appears negatively correlated with network width, which is \textbf{in line with \ref{fac:func}:} an increase in network capacity results in less pressure towards over-generalizing to infrequently seen action values and an ultimately smaller tandem effect. On the other hand, the gap seems to correlate \textit{positively} with network depth. We speculate that this may relate to the finding that deeper networks tend to be biased towards simpler (e.g.~lower rank) solutions, which may suffer from increased over-generalization \citep{huh21low,kumar2020implicit}.

\begin{figure}
    \centering
    \includegraphics[width=\textwidth,trim = {0 0.4cm 0 0.3cm}, clip]{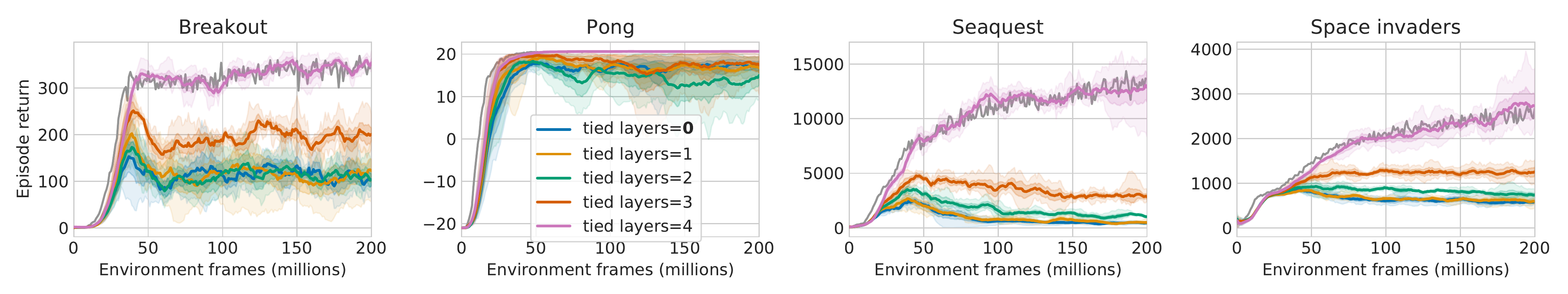}
    \caption{Active vs.~passive performance, with first $k$ of 5 layers of active/passive networks shared.}
    \label{fig:tied_layers}
\end{figure}

Finally, we investigate varying the function class of {\em only the passive network} by sharing the weights of the first $k$ (out of $5$) layers of active and passive networks, while constraining the passive network to only update the remaining top $5-k$ layers, and using the
`representation' at layer $k$ acquired through active learning only. This reduces the `degrees of freedom' of the passive agent, which we hypothesize reduces its potential for divergence.
Indeed, Fig.~\ref{fig:tied_layers} illustrates that passive performance correlates strongly with the number of \textit{tied} layers, with the variant for which only the linear output layer is trained passively performing on par with the active agent. A similar result is obtained in the forked tandem setting, see Appendix Fig.~\ref{fig:fork_tied}. 
This finding provides a strong indirect \textbf{hint towards \ref{fac:func}}: with only part of the network's layers being trained passively, much of its `generalization capacity' is shared between active and passive agents. States that are not aggregated by the shared bottom layers (only affected by active training) have to be `erroneously' aggregated by the remaining top layers of the network for over-generalization to occur. A more thorough investigation of this, exceeding the scope of this paper, may involve attempting to measure (over-)generalization more directly, e.g.~via \textit{gradient interference} \citep{bengio2020interference}.

\section{Applications of the Tandem Setting}

In addition to being valuable for studying the challenges in offline RL,
we propose that the Tandem RL setting provides analytic capabilities that make it a useful
tool in the empirical analysis of general (online) reinforcement learning algorithms.
At its core, the tandem setting aims to decouple learning dynamics from its impact on behavior and
the data distribution, which are inseparably intertwined in the online setting. 
While classic offline RL achieves a similar effect,
as an analytic tool it has the potential downside of typically using a stationary distribution. 
Tandem RL, on the other hand, presents the passive agent with a data distribution which
realistically represents the type of non-stationarity encountered in an online learning process, while
still holding that distribution independent from the learning dynamics being studied.
This allows Tandem RL to be used to study, e.g., the impact of variations in the learning algorithm
on the quality of a learned representation, without having to control for the indirect confounding
effect of a different behavior causing a different data distribution.

While extensive examples of this exceed the scope of this paper,
Appendix \ref{sec:qr} contains a single such experiment, testing
QR-DQN \citep{dabney2018distributional} as a passive learning algorithm (the active agent being a Double-DQN).
This is motivated by the observation of \citet{agarwal20optimistic}, 
that QR-DQN outperforms DQN in the offline setting. QR-DQN indeed appears
to be a nontrivially \textit{different} passive learning algorithm,
significantly better in some games, while curiously worse in others (Fig.~\ref{fig:qr}).

\section{Discussion and Conclusion}

At a high level, our work can be viewed as investigating the issue of (in)compatibility between the data distribution used to train a
function approximator and the data distribution relevant in its evaluation. 
While in supervised learning, \textit{generalization} can be viewed as the problem of transfer from a training to a (given)
test distribution, the fundamental challenge for control in reinforcement learning is that the test distribution is 
\textit{created} by the very outcome of learning itself, the learned policy. 
The various difficulties of learning to act from offline data alone 
throw into focus the role of interactivity in the learning process:
only by continuously interacting with the environment does an agent gradually `unroll'
the very data on which its performance will be evaluated. 

This need not be an obstacle in the case of \textit{exact} (i.e.~tabular) functions:
with sufficient data, extrapolation error can be avoided entirely.
In the case of function approximation however, as small errors compound rapidly into a difference in the underlying
state distribution, significant divergence and, as this and past work demonstrates, ultimately catastrophic under-performance can occur.
Function approximation plays a two-fold role here: (1) being an approximation, it allows deviations in the \textit{outputs};
(2) as the learned quantity, it is (especially in the non-linear case) highly sensitive to variations in the \textit{input distribution}.
When evaluated for control after offline training, these two roles combine in a way that is `unexplored' by the training process: minor {\em output} errors cause a drift in behavior, and thereby a drift in the test distribution.

While related, this challenge is subtly different from the well-known divergence issues of off-policy learning with function approximation, 
demonstrated by Baird's famous counterexample \citep{baird95residual} (see also \citep{tsitsiklis96analysis}) and conceptualized as the Deadly Triad \citep{sutton2018reinforcement,van2018deep}. 
While these depend on bootstrapping as a mechanism to cause a feedback-loop resulting in value
divergence, our results show that the offline learning challenge persists even without bootstrapping,
as small differences in behavior cause a drift in the `test distribution' itself. 
Instead of a \textit{training-time} output drift caused by bootstrapping, the central role is taken by a \textit{test-time} drift of the state distribution caused by the interplay of function approximation and a fixed data distribution (as opposed to dynamically self-generated data). 

Our empirical work highlights the importance of interactivity and `learning from your own mistakes' in learning control.
Starting out as an investigation of the challenges in offline reinforcement learning, it also provides a particular viewpoint on the classical online reinforcement learning case. Heuristic explanations for highly successful deep RL algorithms like DQN, based on intuitions from (e.g.) approximate policy iteration, need to be viewed with caution in light of the apparent hardness of a policy improvement step based on approximate policy evaluation with a function approximator.

Finally, the forked tandem experiments show that even high-performing initializations are not robust to
a collapse of control performance, when trained under their own (\textit{but fixed!}) behavior distribution.
Not just \textit{learning to act}, but even \textit{maintaining performance} appears hard in this setting.
This provides an intuition that we distill into the following working conjecture: 
\textit{The dynamics of deep reinforcement learning for control are unstable on (almost) any \textbf{fixed} data distribution.}

Expanding on the classical on- vs.~off-policy dichotomy,
we propose that indefinitely training on \textit{any fixed} data distribution, 
without strong explicit regularization or additional inductive bias,
gives rise to `exploitation of gaps in the data' by a function approximator, akin to the
over-fitting occurring when over-training on a fixed dataset in supervised learning. Interaction, i.e.~generating at least moderate amounts of one's own experience, appears to be
a powerful, and for the most part \textit{necessary}, regularizer and stabilizer for learning to act, 
by creating a dynamic equilibrium between optimization of a function approximator and its own data-generation process.

\paragraph{Broader impact statement}
This work lies in the realm of foundational RL, contributing to the fundamental understanding and development of RL algorithms, and as such is far removed from
ethical issues and direct societal consequences. 
On the other hand, it highlights the empirical difficulty and limitations 
of offline deep RL for control - increasingly important for practical applications, 
e.g.~robotics, where interactive data is costly, and learning from offline datasets is desirable. 
In this way it complements existing theoretical hardness results in this area and provides additional context
to existing empirical techniques which aim to overcome or circumvent those limitations. 
We believe that an improved understanding of these challenges can play an important role
in creating robust and stable offline learning algorithms whose outputs can be more safely deployed in the real world.

{\bf Acknowledgements}

We would like to thank Hado van Hasselt and Joshua Greaves for feedback on an early draft of this paper, and Zhongwen Xu for an unpublished related piece of work at DeepMind that inspired some of our experiments. We also thank Clare Lyle, David Abel, Diana Borsa, Doina Precup, John Quan, Marc G.~Bellemare, Mark Rowland, Michal Valko, Remi Munos, Rishabh Agarwal, Tom Schaul and Yaroslav Ganin, and many other colleagues at DeepMind and Google Brain for the numerous discussions that helped shape this research.

\bibliographystyle{plainnat}
\bibliography{tandemRL}

\newpage
\appendix

\section{Appendix}

\begin{figure}
    \centering
    \includegraphics[width=\textwidth]{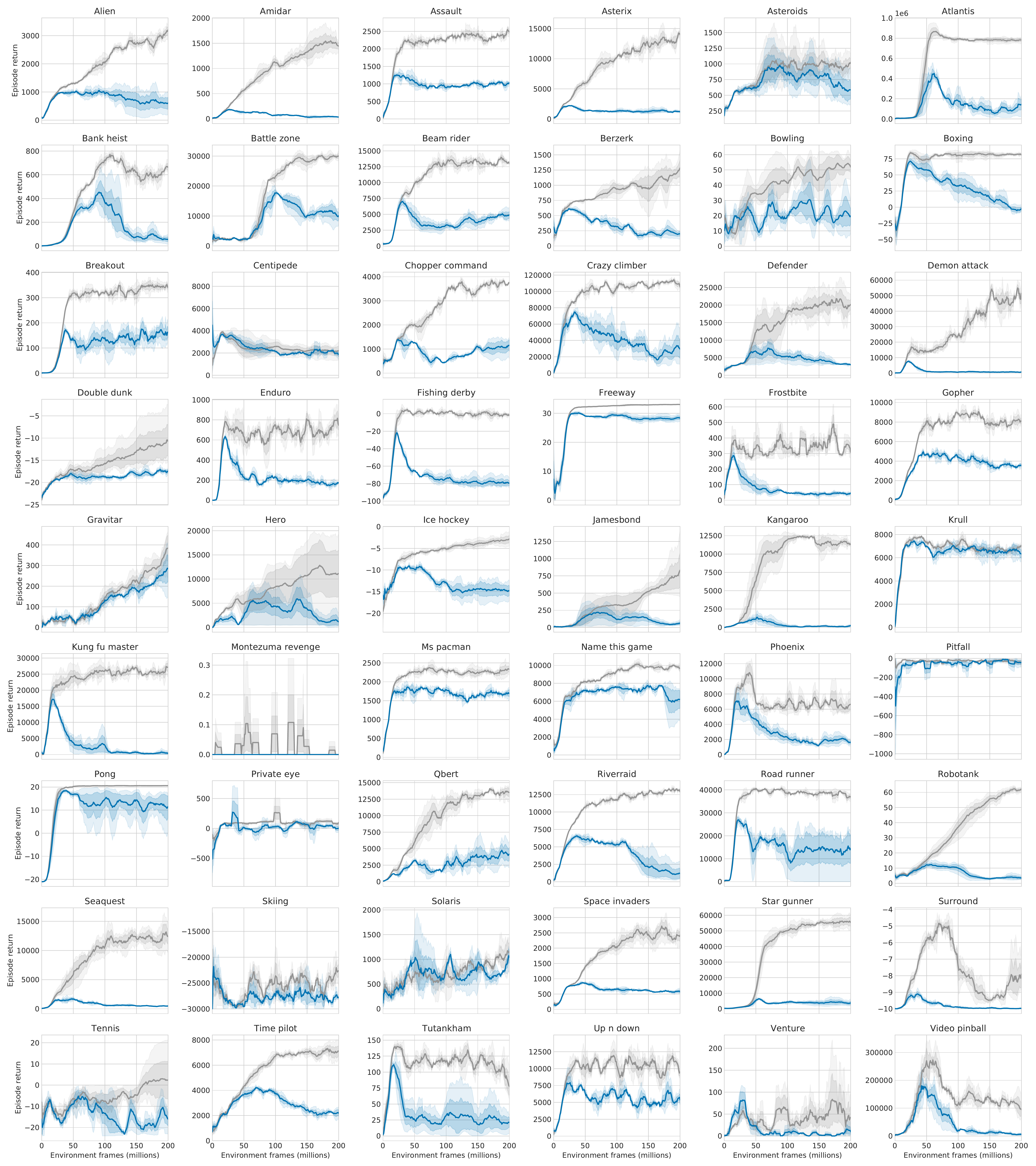}
    \includegraphics[width=0.58\textwidth]{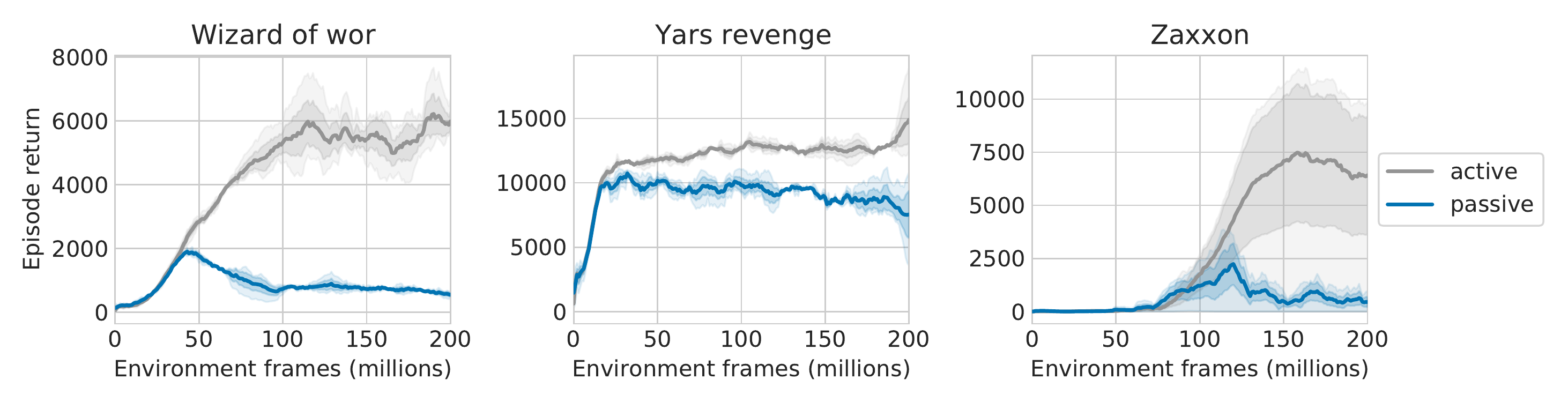}
    \caption{Tandem DQN: active vs.~passive performance across Atari 57 (3 seeds per game).} 
    \label{fig:atari57}
\end{figure}

\begin{figure}
    \centering
    \includegraphics[width=0.5\textwidth]{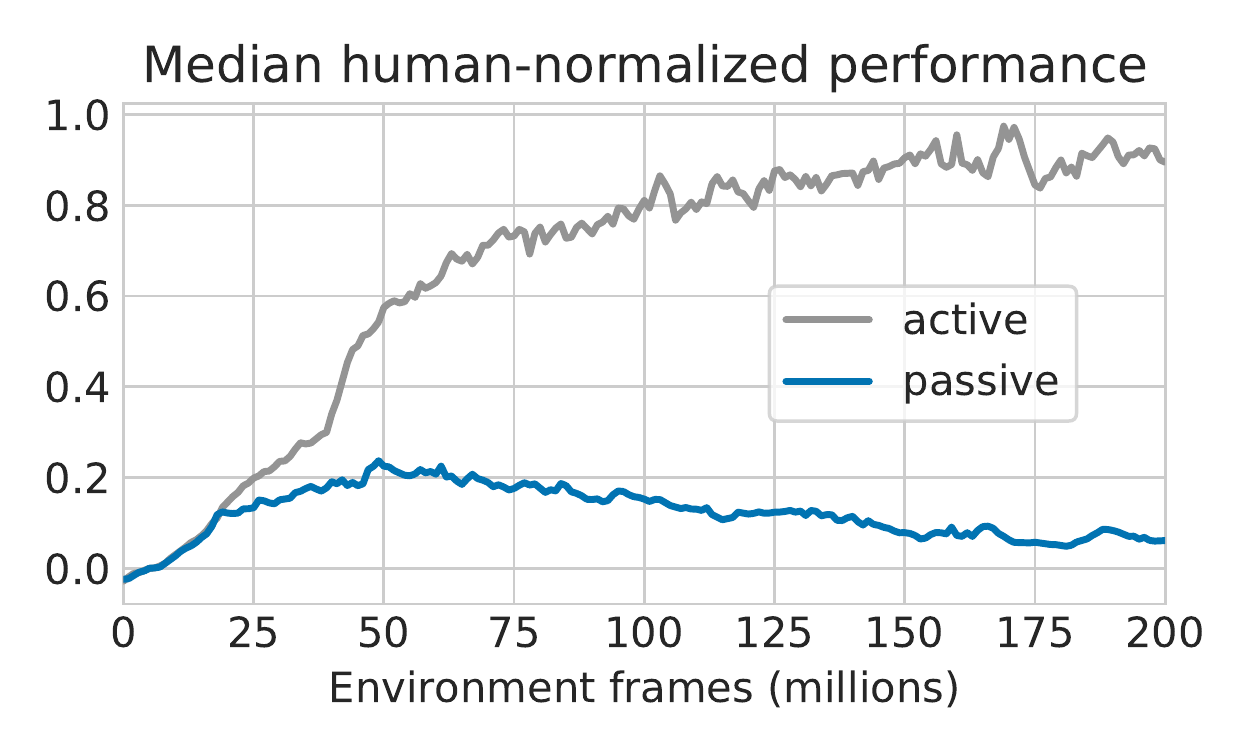}
    \caption{Tandem DQN: Median human-normalized scores over 57 Atari games (3 seeds per game).} 
    \label{fig:atari57_normalized}
\end{figure}

\begin{figure}
    \centering
    \includegraphics[width=\textwidth]{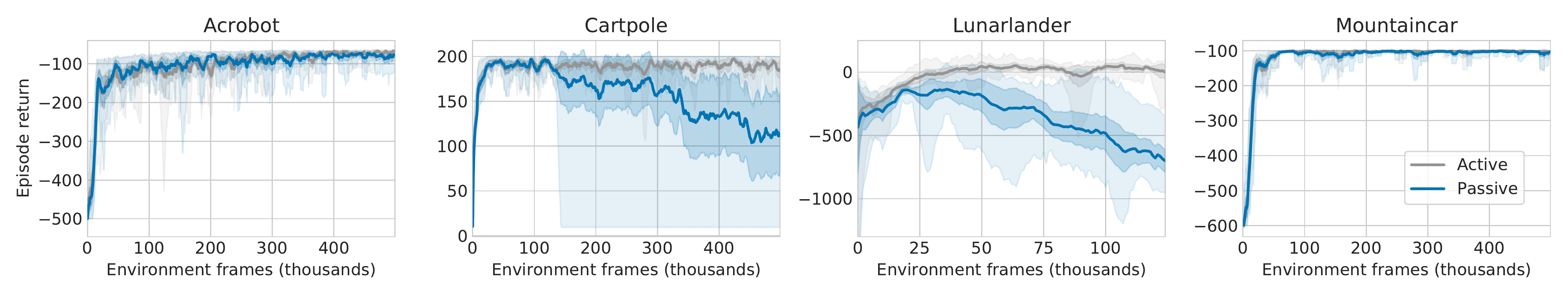}
    \caption{Tandem DQN: Active vs.~passive performance on four selected Classic Control domains.}
    \label{fig:vanilla_tandem_cc}
\end{figure}

\begin{figure}
    \centering
    \includegraphics[width=0.5\textwidth]{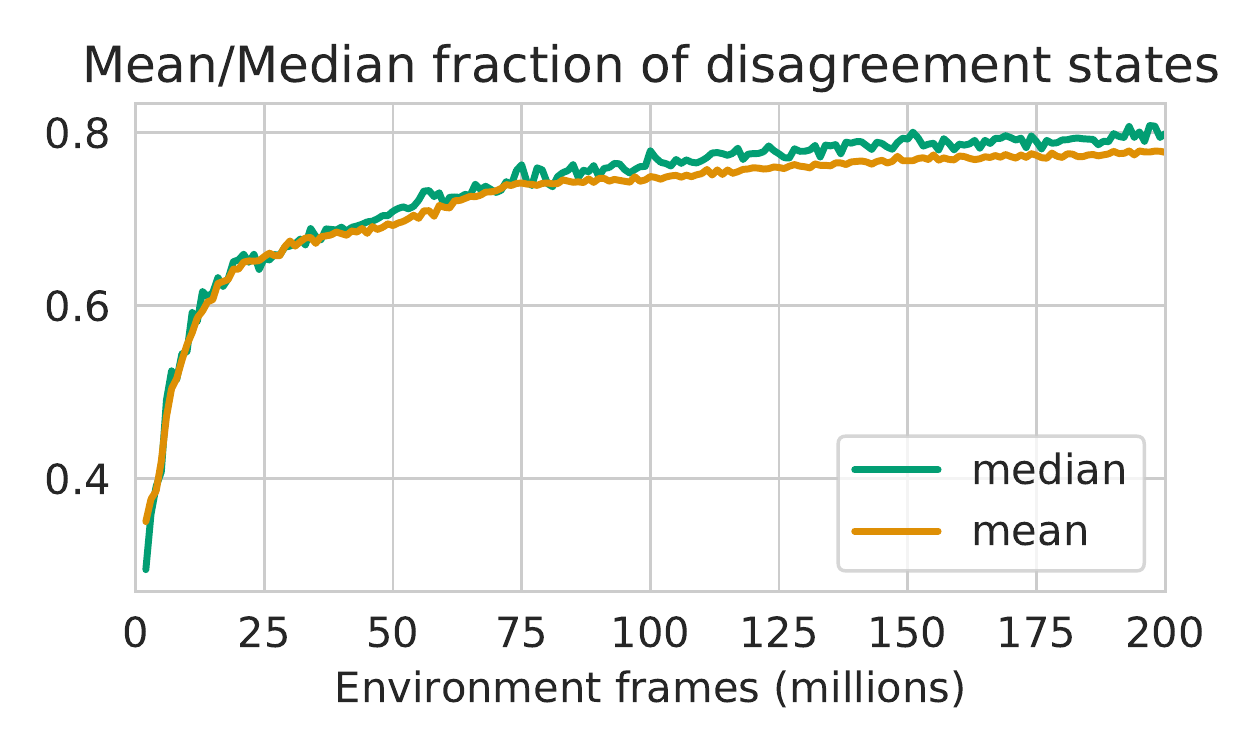}
    \caption{Fraction of states (uniformly sampled from replay buffer) on which active and passive policies disagree,
    i.e.~where $\argmax_a Q_A(s,a) \neq \argmax_a Q_P(s,a)$, mean and median across 57 Atari games (3 seeds per game).} 
    \label{fig:atari57_disagreement}
\end{figure}

\begin{figure}
    \centering
    \includegraphics[width=\textwidth]{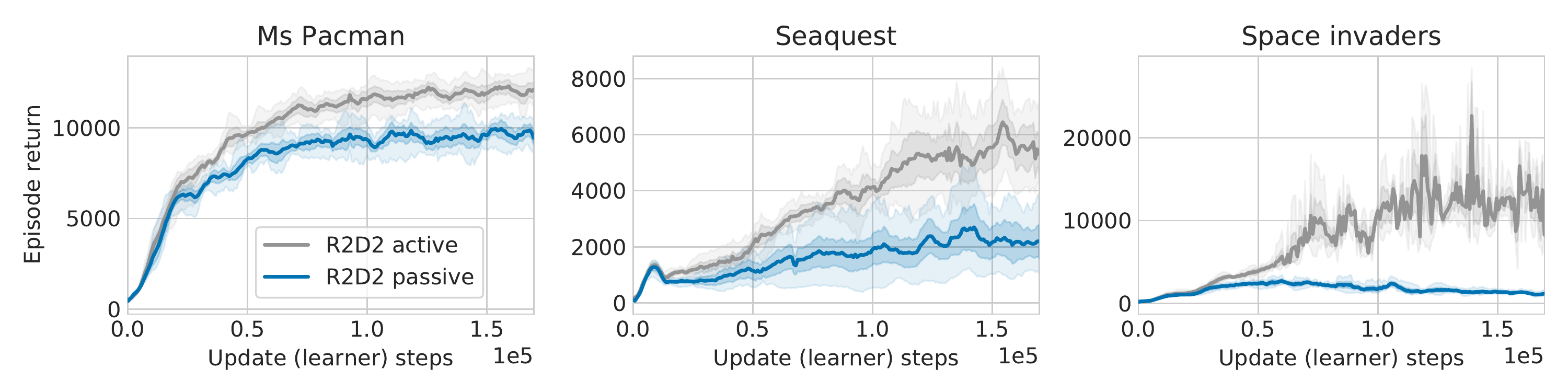}
    \caption{Tandem R2D2: active vs.~passive performance on three Atari domains (3 seeds per game). 
    Note: because of the use of an untuned implementation of R2D2, active results are not directly comparable to those of the published agent \citep{kapturowski2018recurrent}.} 
    \label{fig:r2d2}
\end{figure}

\begin{figure}
    \centering
    \includegraphics[width=\textwidth]{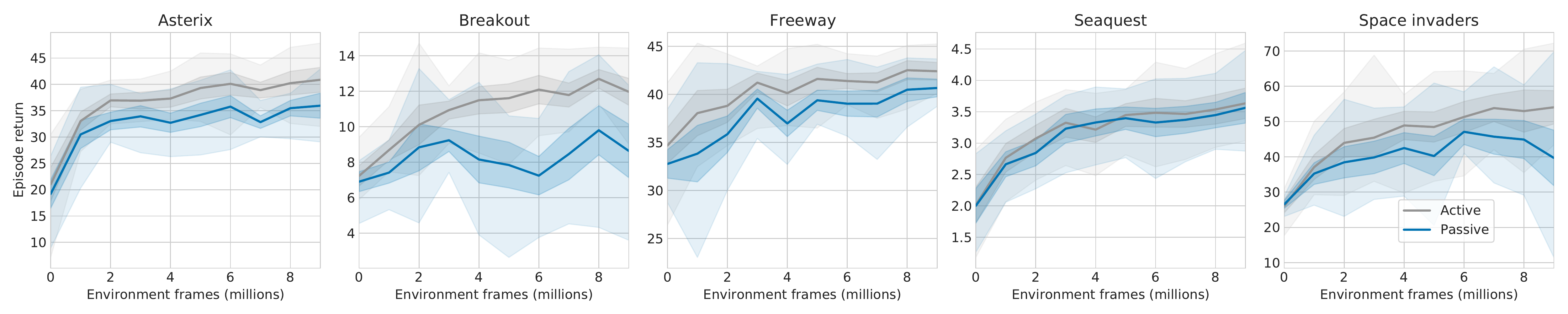}
    \caption{Tandem DQN evaluated on five MinAtar domains.} 
    \label{fig:minatar}
\end{figure}

\begin{figure}
    \centering
    \includegraphics[width=\textwidth]{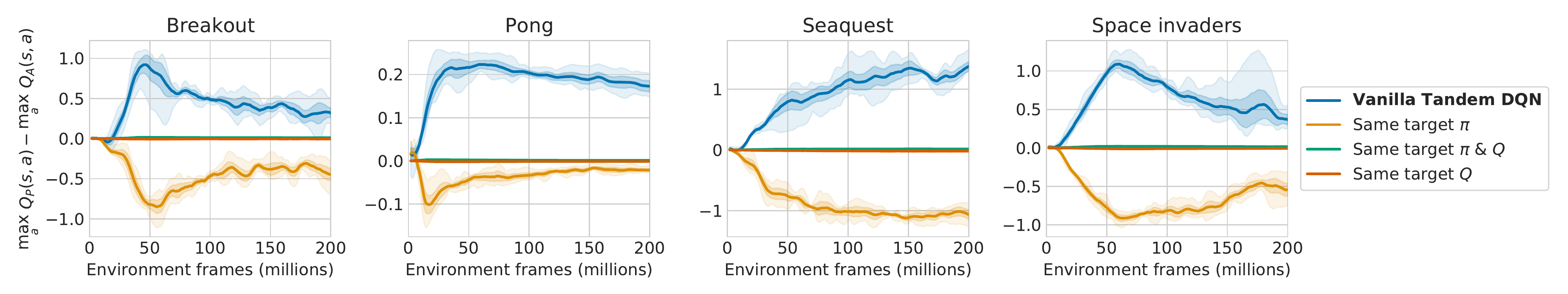}
    \caption{Q-value over-estimation by the passive network compared to the active one in Tandem DQN with varying bootstrap targets. It can be seen that the passive network in the vanilla Tandem DQN setting tends to over-estimate values (compared to the active one), which is almost perfectly mitigated by using the same bootstrap target \textit{values} as the active network, and in fact reversed when using the same target \textit{policy} (but not the same target values). Note that in all four configurations the passive agents substantially under-perform their active counterparts (Fig.~\ref{fig:bootstrapping}), showing that bootstrapping-amplified over-estimation is only part, not the main cause of the tandem effect.}
    \label{fig:bootstrapping_overestimation}
\end{figure}

\begin{figure}
    \centering
    \includegraphics[width=\textwidth]{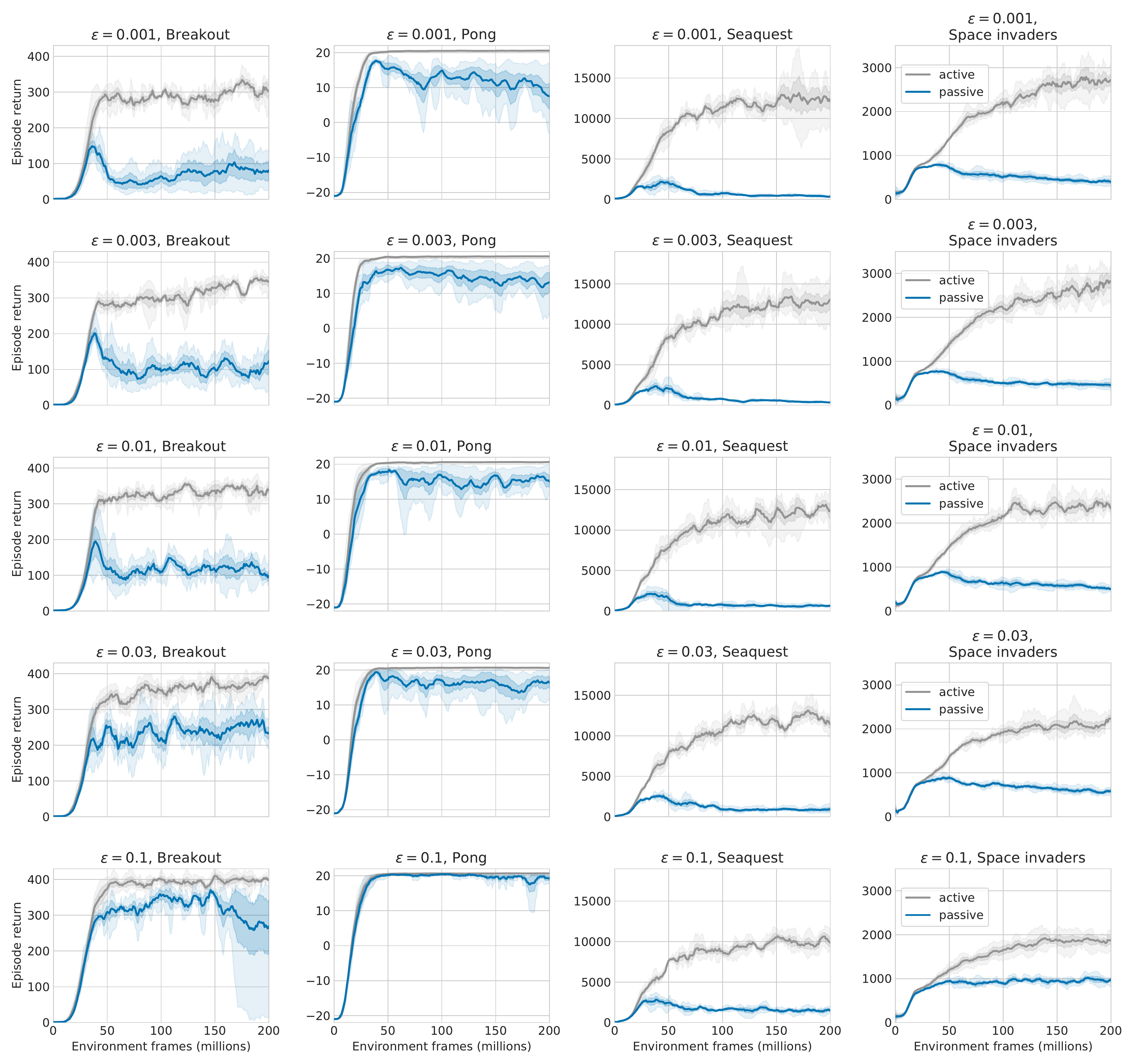}
    \caption{Active vs.~passive performance for varying active $\varepsilon$-greedy behavior policies. Note that here \textit{active} performance varies across settings.}
    \label{fig:eps_return}
\end{figure}

\begin{figure}
    \centering
    \includegraphics[width=\textwidth]{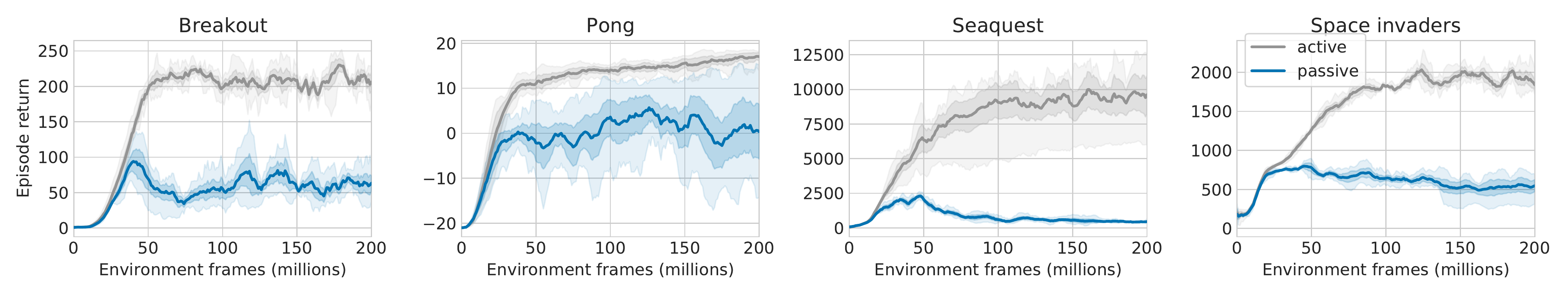}
    \caption{Active vs.~passive performance on Atari with sticky actions \citep{machado2018revisiting}.}
    \label{fig:sticky_actions_return}
\end{figure}

\begin{figure}
    \centering
    \includegraphics[width=\textwidth]{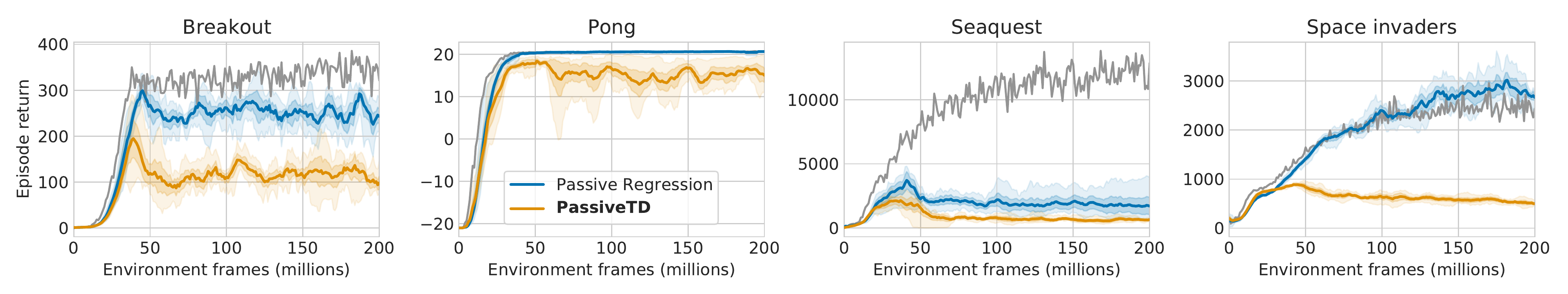}
    \caption{Active vs.~passive performance with regular (TD-based) and regression-based Tandem DQN. The latter regresses all the passive agent's action-values towards the respective outputs of the active agent's network, which can be viewed as network distillation.} 
    \label{fig:regression}
\end{figure}

\begin{figure}
    \centering
    \includegraphics[width=\textwidth]{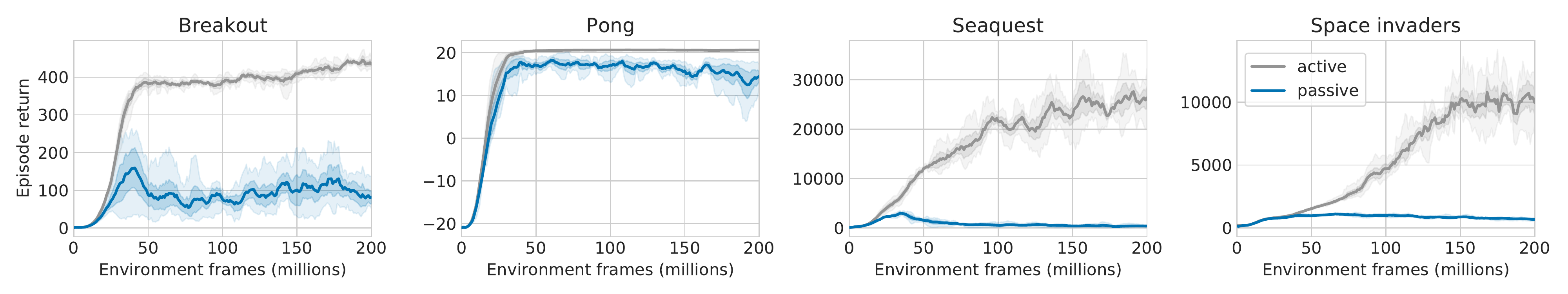}
    \includegraphics[width=\textwidth]{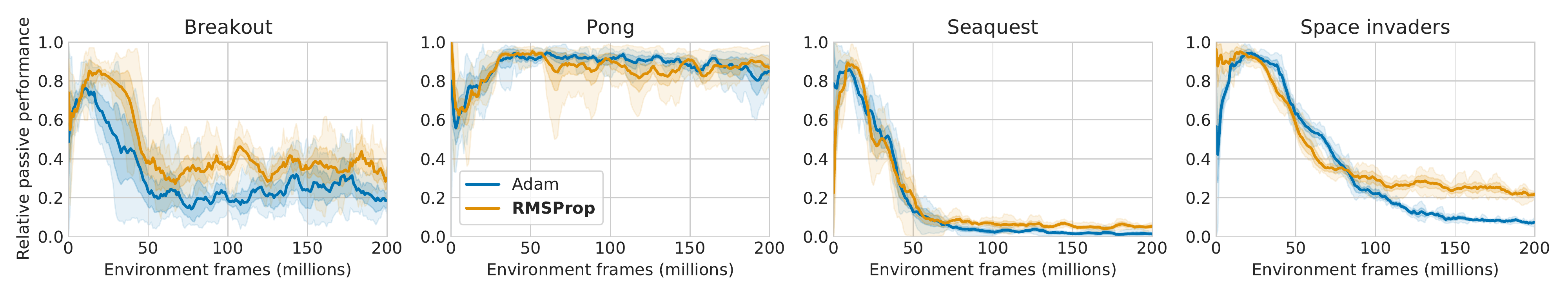}
    \caption{Tandem DQN with the Adam optimizer (instead of RMSProp) for both active and passive network optimization. (\textbf{top}) Adam: Active vs.~passive performance. (\textbf{bottom}) Passive as fraction of active performance Adam vs.~RMSProp.
    While the Adam optimizer improves both the active and passive performance of the Tandem DQN, the relative active-passive gap is not affected strongly.} 
    \label{fig:optim}
\end{figure}

\begin{figure}
    \centering
    \includegraphics[width=\textwidth]{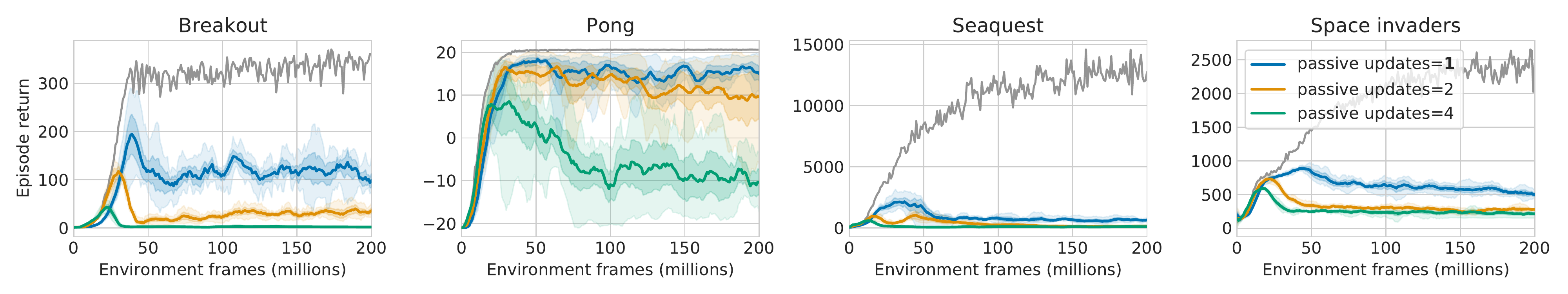}
    \caption{Active vs.~passive performance for varying number of passive updates per active update.}
    \label{fig:replay_ratio}
\end{figure}

\begin{figure}
    \centering
    \includegraphics[width=\textwidth]{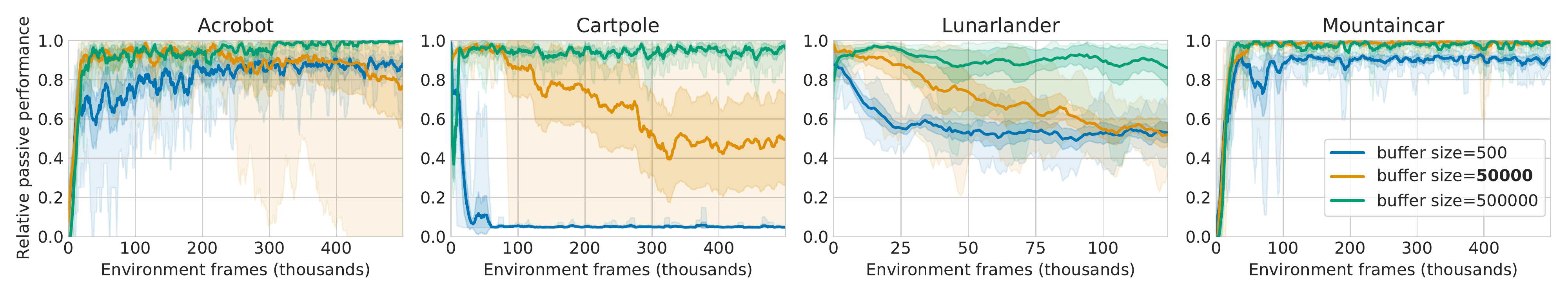}
    \caption{Passive performance as a fraction of active performance when varying the size of the replay buffer used by the passive agent on Classic Control domains.}
    \label{fig:size_cc}
\end{figure}

\subsection{Implementation, Hyperparameters and Evaluation Details}

The implementation of our main agent, Tandem DQN, is based on the Double-DQN \citep{vanhasselt16deep}
agent provided in the DQN Zoo open-source agent collection \citep{dqnzoo2020github}. 
The code uses JAX \citep{jax2018github}, and the Rlax, Haiku and Optax libraries 
\citep{rlax2020github,haiku2020github,optax2020github} for RL losses, neural networks and optimization algorithms, respectively.
All algorithmic hyperparameters correspond to those in the DQN Zoo implementation of Double-DQN.

In the `Tandem' setting, active and passive agents' networks weights are initialized independently.
Agents in this setting are trained in lockstep, i.e.~active and passive agents are updated simultaneously 
from the same batch of sampled replay transitions, with the exception of one experiment in Section \ref{sec:func}, where we study the effect of the number of passive updates relative to active agent updates.

In the `Forked Tandem' setting, only one of the agents is trained at any one time. The active agent trains
(as a regular Double-DQN) up to the time of forking, at which point the passive agent is created as a `fork'
(i.e., with identical network weights) of the active agent. After forking, only the passive agent is trained.
The active agent is used for data generation, either by executing its policy and continuing to fill the replay buffer
(`Fixed Policy' experiment), or by sampling batches from its `frozen' last replay buffer obtained in
the active phase of training (`Fixed Replay' experiment).

The total training budget is kept fixed at 200 iterations in both settings, 
split across `active' and `passive' training phases in the forked tandem setting. 
In all cases, both active and passive agents are evaluated after each training iteration for 500K
environment steps. 
Executed with an NVidia P100 GPU accelerator, each Atari training run takes approximately 4.5 days of wall-clock time.

The majority of our Atari experiments use the regular ALE Atari environment \citep{bellemare13arcade}, using DQN's default
preprocessing, random noop-starts and action-repeats \citep{dqn}, as well as using the reduced action-set (i.e. each game
exposing the subset of Atari's total 18 actions which are relevant to this game). For the `Sticky actions' experiment, 
we use the OpenAI Gym variant of Atari \citep{brockman2016openai} enhanced with sticky actions \citep{machado2018revisiting}.

Unless stated explicitly, all our results are reported as mean episode returns averaged across 5 seeds, 
with light and dark shading indicating $0.5$ standard deviation confidence bounds and min/max bounds (across seeds), respectively. Gray curves always indicate active performance. 

The `relative passive performance' (or `passive performance as fraction of active performance') curves are meant to illustrate the relative (under-)performance of the
passive agent compared to its active counterpart in cases where the active agent's performance varies strongly across configurations.
Denoting $R_a(t)$, $R_p(t)$ the active and passive (undiscounted) episodic returns at iteration $t \in \{0, \ldots, 200\}$, and setting
$m = \min_t \min (R_a(t), R_p(t))$, the relative performance is computed as
\[
 \frac{R_p(t) - m}{R_a(t) - m}
\]
with the value being clipped to lie in $[0, 1]$ and set to $1.0$ whenever $R_a(t) = m$.

For the classic control \citep{brockman2016openai} and MinAtar \citep{young19minatar} 
experiments we used a modified version of the DQN agent from the Dopamine library \citep{castro18dopamine}. 
The modifications made were:
 \begin{itemize}
     \item Double-DQN \citep{vanhasselt16deep} learning updates instead of vanilla DQN
     \item MSE loss instead of Huber loss (as suggested in \citep{obando2020revisiting})
     \item Networks and wrappers for running MinAtar with the Dopamine agents
     \item Tandem training regime (regular and/or forked) instead of regular single-agent training.
  \end{itemize}

Unless stated explicitly, all hyperparameters follow the respective default configurations in the Dopamine library.
Our network architecture for the classic control environments are two fully connected layers of 512 units (each with ReLu activations), followed by a final fully connected layer that yields the Q-values for each action. In Figs.~\ref{fig:netarchCartpole} and \ref{fig:netarchAll} we varied the number of hidden layers and units, where the variation of number of units is applied uniformly across all layers.

The default network architecture for the MinAtar environments is one convolutional layer with 16 features of $3\times 3\times 3$ and stride of $1$ followed by a ReLu activation, whose output is mapped via a fully connected layer to the network's Q-value outputs.

The classic control environments were all run on CPUs; each run took between 20 minutes (\textsc{CartPole}) and 2 hours (\textsc{MountainCar}). The MinAtar environments were all run on NVidia P100 GPUs, each run taking approximately 12 hours to complete. Results for all classic control and MinAtar environments are reported as mean episode returns averaged across 10 seeds, with light and dark shading indicating $0.5$ standard deviation confidence bounds and min/max bounds (across seeds), respectively.

For the R2D2 experiment (Fig.~\ref{fig:r2d2}), an untuned variant of the distributed R2D2 algorithm \citep{kapturowski2018recurrent} was used. Each run used 4 TPUv3 chips for learning and inference, together with a fleet
of approximately 500 CPU-based actor threads for distributed environment interaction, completing a training run of approximately 150K batch updates in about 7 hours wall-clock time.

\subsection{Forked Tandem: Variants}
\label{sec:forked_variants}

Here we present additional experimental variants performed within the Forked Tandem setup.

\paragraph{Varying exploration parameter $\varepsilon$ with fixed policy:} This experiment is an extension of the `Fixed Policy' (Fig.~\ref{fig:forking}) and `The exploration parameter $\varepsilon$' (Fig.~\ref{fig:stochasticity} (top)) experiments. After freezing
the active agent's policy for further data generation, its $\varepsilon$ parameter is set to a different value, to explore the impact
of the resulting policy stochasticity on the ability of the passive learning process to maintain the initial performance level. We note
that because of the fixed active policy, in this case active training performance does not depend on the chosen configuration, and so
absolute passive performance curves are more directly comparable.

Similar to the results in the regular tandem setup, we observe (in Fig.~\ref{fig:eps_return}) that the ability of the passive agent to
maintain the initial performance level is substantially aided by the stochasticity resulting from a higher value of $\varepsilon$,
providing further \textbf{support for the importance of \ref{fac:data}}.

\paragraph{Training process samples (`Groundhog day'):} The forked tandem experiments in Section \ref{sec:distribution} indicate
that data distributions represented by a fixed replay buffer or a stream of data generated by a single fixed policy both show a lack of diversity leading to a catastrophic collapse of an (initially high-performing) agent when trained passively.
The naive expectation that the (unbounded)
stream of data generated by a fixed policy may provide a better state-action coverage than the fixed-size dataset of a single replay buffer (1M transitions) is invalidated by the observation of the fixed-replay training leading to somewhat slower degradation of passive performance. Unsurprisingly in hindsight, the diversity given by the samples stemming from many different policies along a learning trajectory of 1M training steps appears to be significantly higher than that generated by a single $\varepsilon$-greedy policy.

\begin{figure}
    \centering
    \includegraphics[width=\textwidth]{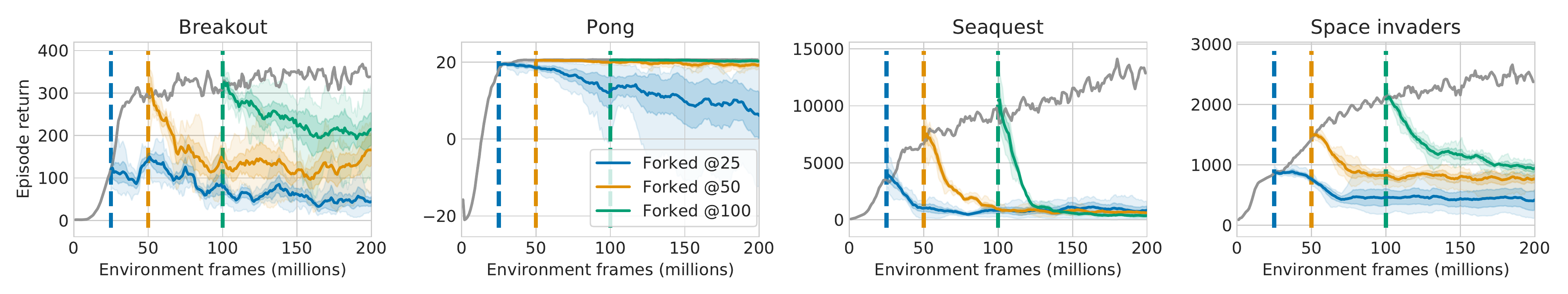}
    \caption{Forked Tandem DQN: `groundhog day' variation, active vs.~passive performance. After forking, the active agent trains for a full iteration (1M environment steps), and is then reset to its initial network parameters at the time of forking, repeatedly for the remaining number of iterations.} 
    \label{fig:ghd}
\end{figure}

To probe this further, we devise an experiment attempting to combine both: instead of freezing the active policy after forking, we continue training it (and filling the replay buffer), however after each iteration of 1M steps, the active network is reset to its parameter values at forking time. Effectively this produces a data stream that can be viewed as producing \textit{samples of the training process} of a single iteration, a variant that we refer to as the `Groundhog day' experiment. This setting combines the multiplicity of data-generating policies with the property of an unbounded dataset being presented to the passive agent. The results are shown in Fig.~\ref{fig:ghd} - indeed we observe that the groundhog day setting improves passive performance over the fixed-policy setting, while not clearly changing the outcome in comparison to the fixed-replay setting.

Overall we observe a general robustness of the tandem effect with respect to minor experimental variations in the forked tandem settings.

\begin{figure}
    \centering
    \includegraphics[width=\textwidth]{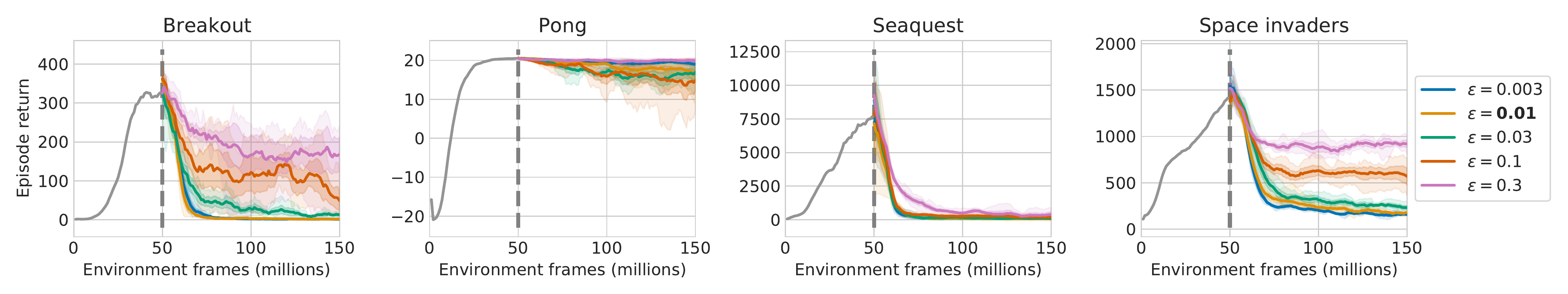}
    \caption{Forked Tandem DQN: After 50 iterations of regular active training, the active value function is frozen and used to continuously
    generate data for the passive agents' training by executing an $\varepsilon$-greedy policy with a given value of $\varepsilon$.}
    \label{fig:fork_eps_return}
\end{figure}

\subsection{Forked Tandem: Policy Evaluation} \label{sec:eval}

\begin{figure}
    \centering
    \includegraphics[width=\textwidth]{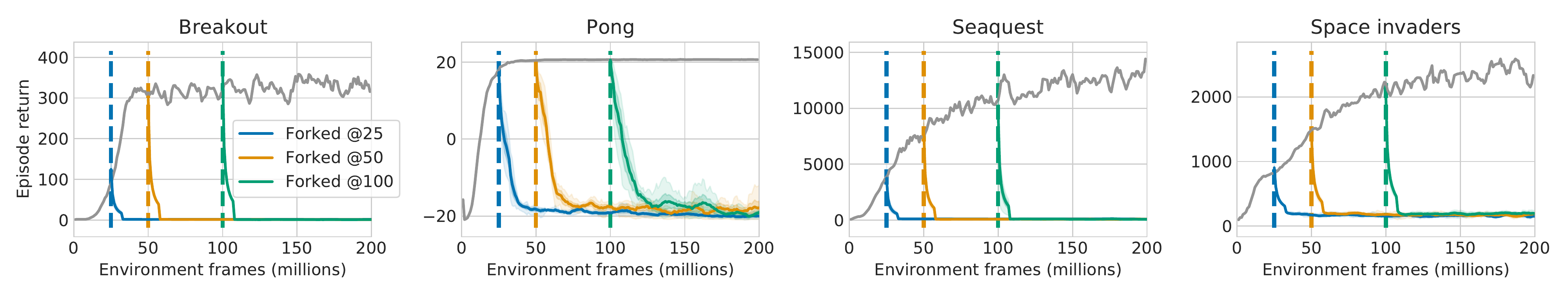}
    \includegraphics[width=\textwidth]{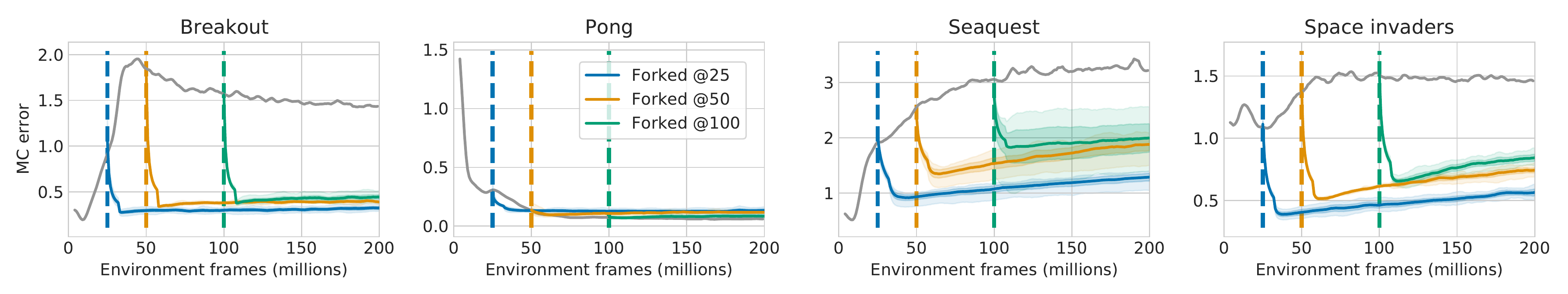}
    \caption{Forked Tandem DQN: Policy evaluation with Monte-Carlo return regression. 
    (\textbf{top}) Active vs.~passive control performance. (\textbf{bottom}) Average absolute Monte-Carlo error. The Monte-Carlo error is minimized effectively by the passive training, while the control performance of the resulting $\varepsilon$-greedy policy collapses completely.} 
    \label{fig:mc}
\end{figure}

Among the most striking findings in our work are the forked tandem findings 
in Sections \ref{sec:distribution} and \ref{sec:forked_variants},
demonstrating a catastrophic collapse of performance when passively training from a fixed data distribution, even when the starting
point for the passive policy is the very same high-performing policy generating the data. This leads to the question whether the process
of Q-learning itself is to blame for this failure mode, 
e.g.~via the well-known statistical over-estimation bias introduced by the $\max$ operator \citep{hasselt2010}. 
To test this, we perform two variants of the forked tandem experiment with SARSA \citep{rummery1994line} and (purely supervised) Monte-Carlo return regression based policy evaluation instead of Q-learning as the passive learning algorithm.
(We note that while SARSA evaluation of an $\varepsilon$-greedy policy can still exhibit over-estimation bias, this is not the case
for Monte-Carlo return regression.)

As can be seen in Figs.~\ref{fig:sarsa} and \ref{fig:mc}(top), even in this on-policy policy evaluation setting, the resulting control performance catastrophically collapses after a short length of training. We also observe that this is not due to a failing of the evaluation: Fig.~\ref{fig:mc}(bottom) shows effective minimization of the Monte-Carlo error, indicating that the control failure is due to extrapolation error (and in particular, over-estimation) on infrequent (under the active policy) state-action pairs.

These findings provide strong \textbf{support for the role of \ref{fac:data} and \ref{fac:func}} while \textbf{weakening that of \ref{fac:bootstrapping}}: In contrast to the well-known `Deadly Triad' phenomenon \citep{sutton2018reinforcement,van2018deep}, the tandem effect occurs without the amplifying mechanism of bootstrapping or the statistical over-estimation caused by the $\max$ operator, solely due to erroneous extrapolation by the function approximator to
state-action pairs which are under-represented in the given training data distribution. 

\begin{figure}
    \centering
    \includegraphics[width=\textwidth]{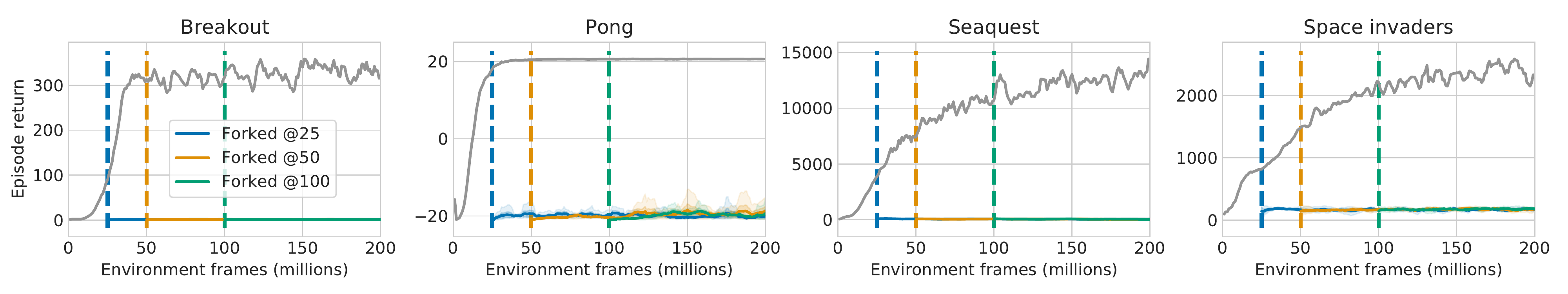}
    \includegraphics[width=\textwidth]{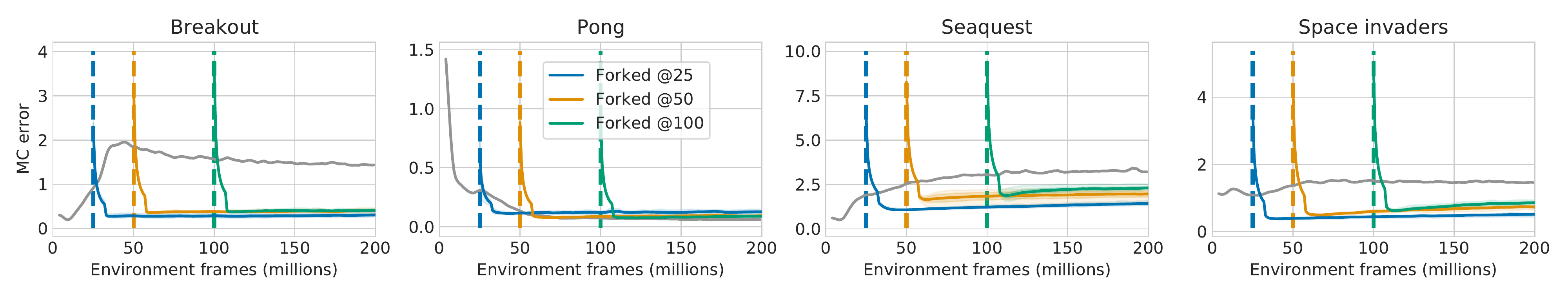}
    \caption{Forked Tandem DQN: Policy evaluation with Monte-Carlo return regression, passive network initialized independently
    of the active network at forking time.
    Top: Active vs.~passive control performance. Bottom: Average absolute Monte-Carlo error. While Monte-Carlo error is minimized effectively by the passive training, this does not result in above-random control performance of 
    the resulting $\varepsilon$-greedy policy.} 
    \label{fig:mc_fresh}
\end{figure}

We have so far presented the equal network weights of active and passive networks at forking time as a strength of the forked tandem setting, following the intuition that an initialization by the high-performing policy should be advantageous for maintaining performance
in these experiments. A plausible counter-argument could be that the representation (i.e., the network weights) learned by the active agent \textit{in service of control} could be a poor, over-specialized starting point for policy evaluation. To verify that this is not a major factor, we also perform the above Monte-Carlo evaluation experiment with the passive network freshly re-initialized at the beginning of passive training. As shown in Fig.~\ref{fig:mc_fresh}, while a fresh initialization of the passive network indeed allows it to similarly effectively minimize Monte-Carlo error, its control performance here never exceeds random performance levels, further \textbf{connecting the tandem effect to \ref{fac:data} and \ref{fac:func}.}

We remark that the demonstrated control performance failure of approximate policy evaluation casts a shadow over the concept of
approximate policy iteration, or the application of this concept in heuristic explanations of the function of empirically successful  algorithms like DQN \citep{dqn}. A successful greedy improvement step on an \textit{approximately evaluated} policy appears implausible
given the brittleness of approximate policy evaluation \textit{even in the nominally best-case scenario of an on-policy data distribution}. 

Another view point emerging from these results is that the classic category of `on-policy data' appears less relevant in this context: an appropriate data distribution for robust approximate evaluation targeting \textit{control} seems to require a data distribution sufficiently overlapping with the (hypothetical, in practice unavailable) behavior distribution of the \textit{resulting policy} rather than the original evaluated policy.

\subsection{Additional Experimental Results on the Role of Function Approximation}

Here we present several extra experiments, complementing the results from the end of Section \ref{sec:func}.

\begin{figure}
    \centering
    \includegraphics[width=\textwidth]{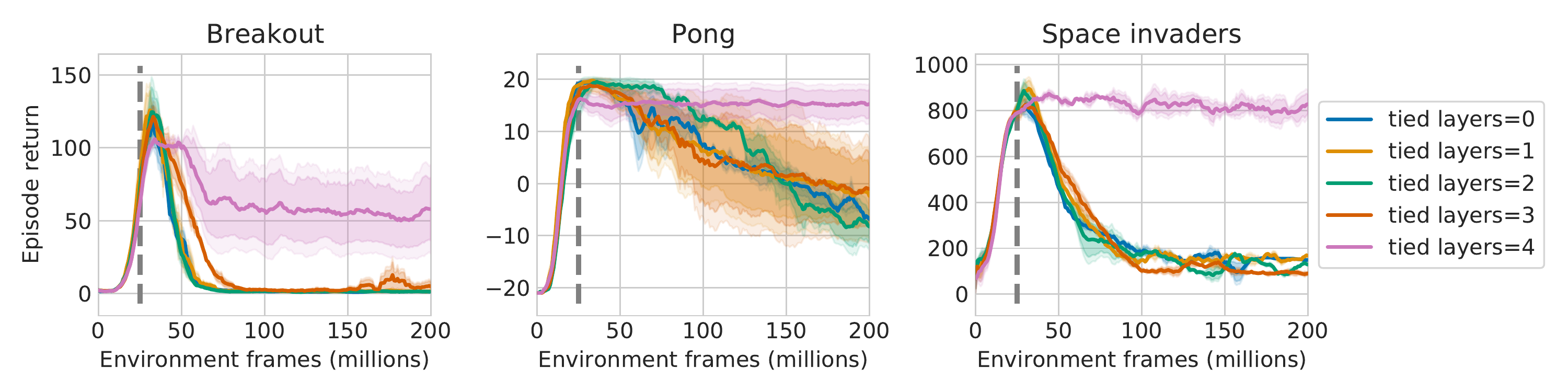}
    \includegraphics[width=\textwidth]{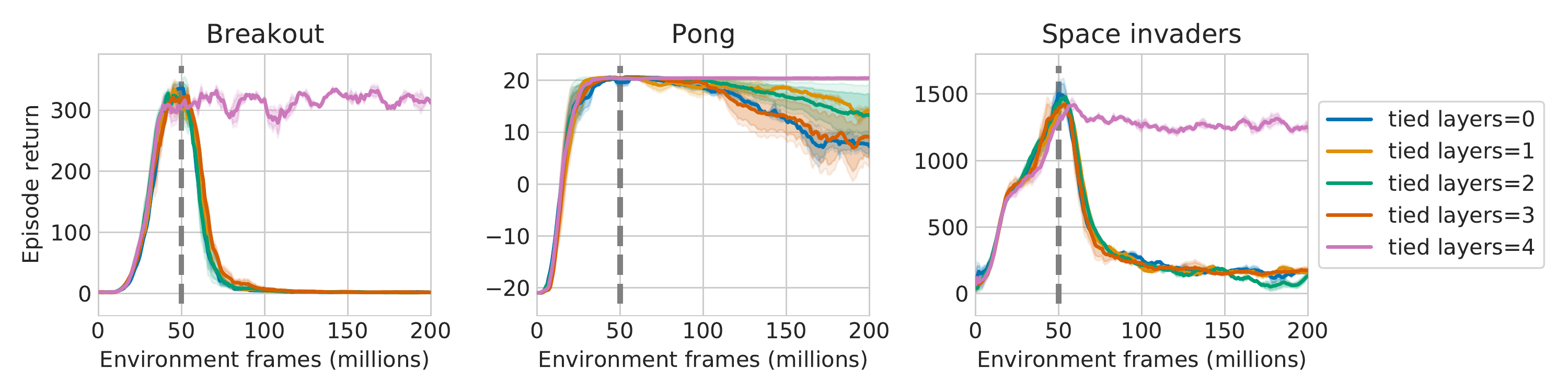}
    \includegraphics[width=\textwidth]{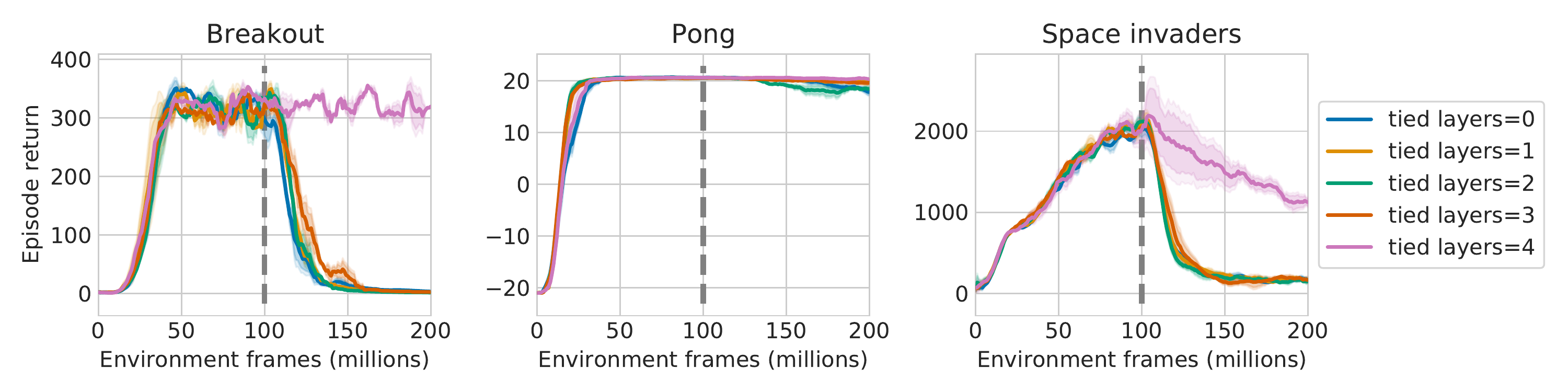}
    \caption{Forked Tandem DQN: passive performance after forking, with the first $k$ (out of $5$) 
    layers of active and passive agent networks shared, i.e.~only trained by the active agent, and fixed during the passive training phase.} 
    \label{fig:fork_tied}
\end{figure}

The first set of results, shown in Fig.~\ref{fig:fork_tied}, concerns the passive performance in the forked tandem setup, when the first (bottom) neural network layers are shared between active and passive agents. These bottom layers are only trained during the active training phase before forking, and are frozen after that. Similar to the corresponding experiment in the regular tandem setting (Fig.~\ref{fig:tied_layers}), we observe that the passive agent's ability to maintain its initially high performance strongly correlates with the number of \textit{shared}, i.e.~not passively trained layers. The difference between the passive agent training a `deep' vs.~a `linear' network (the latter corresponds to all but one of the network layers being frozen) appears stark: the tandem effect is almost equally catastrophic in all configurations except for the linear one, where it appears to be strongly reduced. While a more thorough investigation remains to future work, we remark that overall this finding appears to \textbf{supports the importance of \ref{fac:func}}, in that intuitively over-extrapolation can be expected to become more problematic when passive training is applied to a larger function class (deeper part of the network).

\begin{figure}
    \centering
    \includegraphics[width=\textwidth]{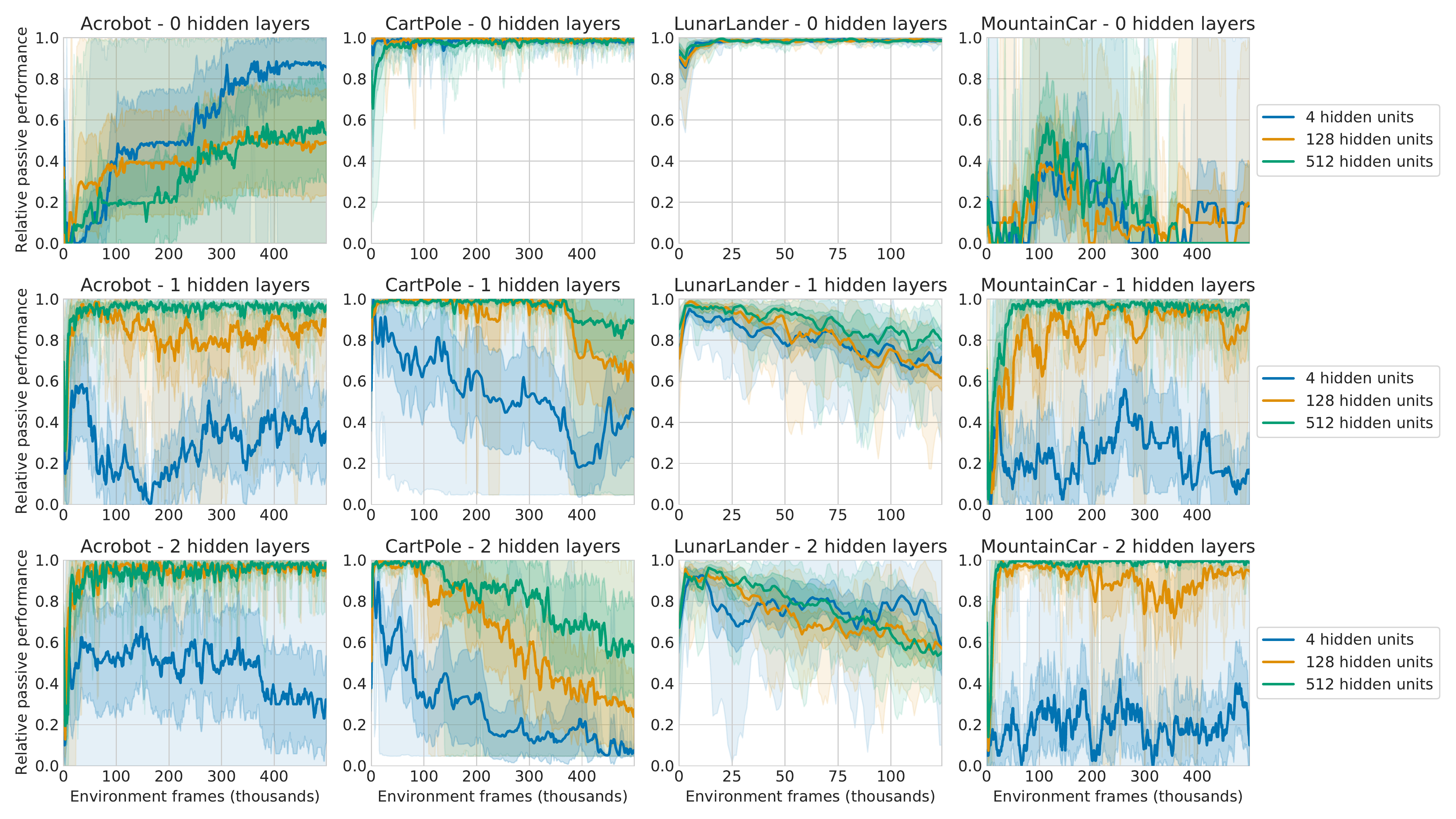}
    \caption{Tandem DQN: Passive performance as a fraction of active performance when varying network architecture in classic control games. Here network architecture varies for both active and passive agent, so active performance is also affected
    by the configuration.} 
    \label{fig:netarchAll}
\end{figure}

The next experiment, shown in Fig.~\ref{fig:netarchAll} and extending the results from Fig.~\ref{fig:netarchCartpole}, investigates the impact of network architecture more generally, by varying width and depth of both active and passive agents' networks. Since changes in Atari-DQN network architecture tend to require expensive re-tuning of various hyperparameters, we chose to perform these experiments on the smaller Classic Control domains, where such changes tend to be
more straightforward. Nevertheless active performance in these domains does depend on the chosen network configuration, so that
we report relative performance as the more informative quantity. The findings across four domains mostly appear to echo those on \textsc{CartPole} described in Section \ref{sec:func}, showing a positive correlation of (relative) passive performance with network \textit{depth}, and a negative correlation with its \textit{width}. Again, a more detailed investigation of the causes for this exceeds the scope of this paper and is left to future work.

\subsection{Applications of Tandem RL: Passive QR-DQN}
\label{sec:qr}

\begin{figure}
    \centering
    \includegraphics[width=\textwidth]{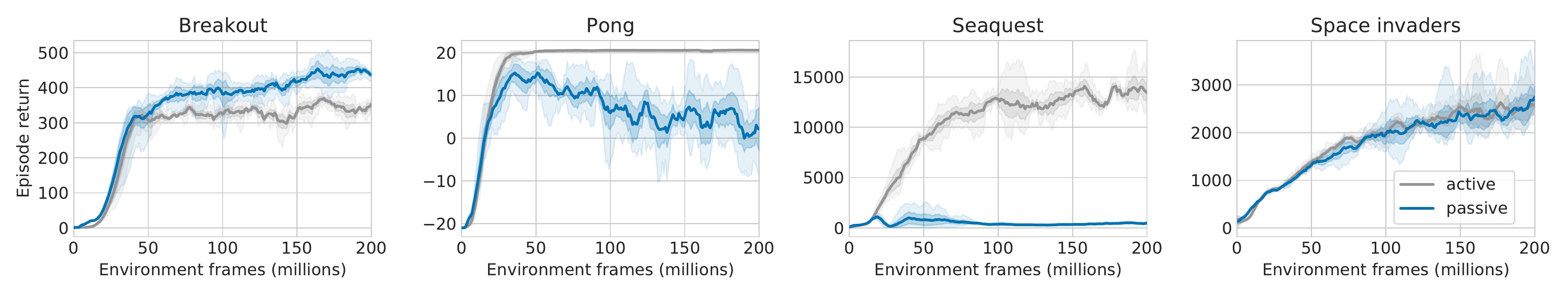}
    \caption{QR-DQN as a passive learning algorithm, in tandem with a Double-DQN active agent: active vs.~passive performance.} 
    \label{fig:qr}
\end{figure}

Here we provide an example application for the Tandem RL setting as an analytic tool for the study of learning algorithms
in isolation from the confounding factor of behavior. As observed in \citep{agarwal20optimistic}, the QR-DQN algorithm \citep{dabney2018distributional} can be preferable to DQN in the offline setting, motivating our attempt to use
it as a passive agent, coupled with a regular Double-DQN active agent. As shown in Fig.~\ref{fig:qr}, QR-DQN indeed provides a somewhat different passive performance profile when compared to the regular Double-DQN tandem, albeit not a clearly better one. While perfectly matching active performance in one game (\textsc{Space Invaders}) and even \textit{out-performing the active agent} in another (\textsc{Breakout}), it also shows exacerbated under-performance or instability in the other two domains. A fine-grained diagnosis of the causes of this are left to future work. 

We note that any difference in performance between the DQN and QR-DQN algorithms as passive agents reflects directly on their
properties as learning algorithms, i.e.~their respective abilities to extract information about an appropriate control policy from
observational data, while separating out any influence their learning dynamics may have on (transient) behavior and data generation. We believe that Tandem RL can become a valuable analytic tool for targeted empirical studies of such properties of learning algorithms. 

\end{document}